\documentclass[11pt]{article}

\usepackage[final]{acl}

\usepackage{times}
\usepackage{latexsym}

\usepackage[T1]{fontenc}

\usepackage[utf8]{inputenc}

\usepackage{microtype}
\usepackage{amsmath}
\usepackage{amssymb}
\usepackage{enumitem}
\usepackage{kotex}
\usepackage{booktabs}
\usepackage{multirow}
\usepackage[most]{tcolorbox}
\usepackage{tabularx}
\usepackage[table]{xcolor}
\usepackage{makecell}
\usepackage{inconsolata}
\usepackage{url}

\usepackage{graphicx}
\definecolor{lowSatBlue}{RGB}{90, 113, 132}
\definecolor{lightBackground}{RGB}{248, 249, 250}

\title{Citation-Closure Retrieval and Per-Rule Attribution for Real-World Regulatory Compliance Question Answering}

\author{
 \textbf{Yeong-Joon Ju\textsuperscript{1}},
 \textbf{Seong-Whan Lee\textsuperscript{1}}
\\
 \textsuperscript{1}Department of Artificial Intelligence, Korea University
\\
 \texttt{\{yj\_ju, sw.lee\}@korea.ac.kr}
}

\begin{document}
\maketitle

\begin{abstract}

Deploying Large Language Models (LLMs) for regulatory compliance demands rigorous traceability via comprehensive citations across multi-tiered authority structures. Unlike traditional multi-hop or legal QA, this task requires structured procedural lookups and evidence-set closure rather than entity resolution or case-law reasoning. Existing RAG systems struggle here due to flattened citation edges, fragmented retrieval expansions, and fragile post-hoc attribution. We formalize Regulatory Compliance QA with RegOps-Bench, a novel benchmark featuring an Operational Knowledge Graph derived from complex national R\&D regulations. To address these bottlenecks, we propose RefWalk, a unified framework driven by a shared topic anchor. RefWalk traverses cross-document citations, fuses multi-view candidates via max-based aggregation, and enforces per-rule attribution to explicitly map claims to sources. We establish a strong baseline with substantial improvements in retrieval recall and citation accuracy. Finally, a contrastive evaluation on a U.S. health compliance dataset (HIPAA) reveals that existing systems exhibit saturation on flat-structure rules, underscoring the need for RegOps-Bench. Our code is available at \url{https://github.com/yeongjoonJu/RefWalk}.

\end{abstract}

\section{Introduction}

Regulated organizations operate under strict, layered frameworks of statutes, enforcement decrees, ministerial rules, notices, and operational manuals. Since non-compliance carries severe risks, compliance staff routinely navigate complex procedural questions whose grounds span these multiple authority tiers~\cite{arner2018regtech}. While Large Language Models (LLMs) offer a promising avenue to alleviate this burden, deploying them to assist with such inquiries requires rigorous traceability~\cite{ariai2025natural,liu2023evaluating}. To establish a verifiable audit trail, models should quote controlling articles and explicitly map every generated claim back to its source.

In this paper, we formalize this objective as Regulatory Compliance Question Answering (QA), where a model addresses a practitioner's query by retrieving the exhaustive set of governing articles and detailing the precise claims derived from each rule. As modern regulatory systems evolve into highly complex cross-reference networks across multiple documents~\cite{katz2020complex,ruhl2015measuring}, this task structurally diverges from both standard multi-hop and legal QA paradigms. Unlike standard multi-hop QA paradigms focusing on entity resolution~\cite{yang-etal-2018-hotpotqa, trivedi-etal-2022-musique}, hop transitions in this QA task follow typed citation rules, and search termination relies on complete evidence-set closure rather than finding a single final entity. Furthermore, while traditional legal QA tasks~\cite{guha2023legalbench, li2024lexeval, yang2026lawthinker} largely center on judicial interpretation and case-law reasoning, recent regulatory NLP has expanded into procedural clause retrieval~\cite{louis-spanakis-2022-statutory} and corporate sustainability extraction~\cite{ali2025sustainableqa}. However, these existing benchmarks predominantly operate on flat rule structures or isolated reports, failing to capture the mechanics of hierarchical compliance. In contrast, Regulatory Compliance QA demands navigating multi-tiered, cross-document delegations to achieve deterministic evidence-set closure.



These rigorous demands expose two critical failure modes in existing RAG systems. The first is the inability to execute precise structural retrieval. Current graph-based RAG approaches~\cite{edge2024local, guo-etal-2025-lightrag, gutierrez2025from, ma2025thinkongraph, peng2025graph} typically flatten explicit regulatory citations into generic entity-relation edges, stripping away the semantic distinctions necessary for procedural navigation. Because they rely on global signal propagation across surface-level entity overlaps, standard entity-centric graphs inevitably fail to resolve the rigid, multi-tiered delegation pathways inherent in regulatory networks. Moreover, because regulatory queries often omit specific situational constraints, systems must exhaustively identify all potential conditional branches and legal exceptions. Attempts to resolve this via query expansion or decomposition~\cite{gao2023precise, rackauckas2024rag, wang2023query2doc, petcu-etal-2026-query} typically rely on surface-level paraphrasing, produce disjoint sub-queries that fragment rather than unify the required evidence set.

The second major limitation emerges during answer generation, where attribution is typically treated as a post-hoc afterthought~\cite{saxena2025generationtime}. In regulatory compliance, the risk of a hallucinated citation outweighs the benefit of a broad response. However, rather than enforcing a schema-level binding between individual claims and their sources, generative models generally append citations as free-form footers, leading to systemic attribution failures~\cite{bohnet2022attributed, liu2023evaluating, hou2025clerc}. Ultimately, this lack of structural binding between claims and their sources fails to provide the rigorous traceability required, leaving a significant gap in the reliability of regulatory AI.

To systematically diagnose these limitations, we introduce RegOps-Bench, an evaluation framework serving as a testbed for multi-tier procedural navigation, instantiated via a highly structured Korean national R\&D regulatory corpus. This corpus features a five-tiered authority structure that exemplifies the nested complexities inherent in real-world administrative regulations. We model this corpus into an Operational Knowledge Graph (OKG) and construct 231 high-quality QA pairs grounded in real-world inquiries from the Institute of Information \& Communications Technology Planning \& Evaluation (IITP). By expanding these inquiries using a novel axis-decoupling principle, we independently control the substantive intent of a query and the structural complexity of its required reference set, spanning from straightforward lookups to exception-heavy procedural branching. This orthogonal design precisely isolates whether a system fails due to navigating a dense regulatory hierarchy or resolving complex procedural logic.

To overcome the identified retrieval and generation bottlenecks, we propose RefWalk, a structural traversal framework that navigates regulatory citation pathways guided by a shared topic anchor. RefWalk mitigates structural retrieval failures by exploring the OKG through three distinct semantic views, restricting hop expansion strictly to cross-document citation edges to eliminate internal containment noise. To preserve the specialized signals required across different difficulty tiers, candidates are fused using Reciprocal Rank MAX (RRM) rather than standard sum-based aggregation~\cite{cormack2009reciprocal}, which otherwise dilutes crucial specialist cues. During generation, RefWalk tackles systemic attribution failures by injecting the same anchor alongside a per-rule schema. Instead of appending citations as a post-hoc afterthought, we structure the model to generate claims directly as attributes of their governing rules. This approach inherently mitigates attribution hallucinations and binds generation to its source, advancing the traceability required for professional practice. Finally, we demonstrate the broader applicability of this evaluation framework and RefWalk by validating both on a HIPAA-derived dataset.

In summary, our main contributions are threefold. First, we formalize the task of Regulatory Compliance QA and release RegOps-Bench, the first benchmark for deterministic traversal of multi-tier regulatory hierarchies. Second, we propose RefWalk, a unified RAG framework that navigates explicit structural delegations, establishing a robust baseline for complex procedural lookups. Third, we outperform state-of-the-art RAG methods on complex cross-reference tasks. Furthermore, ablation studies confirm the impact of our schema-level binding, while contrastive experiments on HIPAA validate the necessity of multi-tiered evaluation.


\section{Related Work}

\subsection{Regulatory NLP and Structural Retrieval}

Legal NLP has increasingly shifted from coarse classification~\cite{chalkidis2020legal,guha2023legalbench} toward reasoning- and retrieval-intensive evaluation~\cite{yang2026lawthinker,hou2025clerc}. Within this space, regulatory compliance presents unique challenges distinct from case-law analogy, requiring the navigation of complex, multi-document networks~\cite{katz2020complex,ruhl2015measuring,sleimi2018automated} to meet the demands of RegTech~\cite{arner2018regtech}. While De Jure~\cite{guliani2026jure} focuses on structuring raw regulations into rule sets, we leverage its evaluation pipeline for our HIPAA data generation. However, existing benchmarks still fail to formalize QA across layered, cross-document procedural chains.

Standard multi-hop QA~\cite{yang-etal-2018-hotpotqa,trivedi-etal-2022-musique,ho-etal-2020-constructing} and graph-based RAG~\cite{edge2024local,guo-etal-2025-lightrag,gutierrez2025from,ma2025thinkongraph,wang2025pike} typically attempt complex retrieval by propagating through entity-centric relations. Similarly, query decomposition~\cite{gao2023precise,wang2023query2doc,rackauckas2024rag,petcu-etal-2026-query,khot2023decomposed,trivedi-etal-2023-interleaving,jiang-etal-2023-active} divides queries into independent sub-facts, while representation-based retrieval methods~\cite{lee2025nvembed,ju-etal-2025-mire,ju2025generator} improve semantic matching.

\subsection{Attributed Generation and Citation Faithfulness}

Verifiable grounding is widely recognized as a strict deployment prerequisite for legal AI~\cite{ariai2025natural,hou2025clerc}. Consequently, citation faithfulness has emerged as a central evaluation axis, driving the development of specialized attribution benchmarks~\cite{bohnet2022attributed,rashkin2023measuring,liu2023evaluating,gao-etal-2023-enabling} and methods that fold retrieval or revision decisions into the generation policy~\cite{asai2024selfrag,gao-etal-2023-rarr}. Despite these advances, audits of off-the-shelf LLMs and domain-specific legal systems~\cite{liu2023evaluating,hou2025clerc} consistently reveal that a substantial fraction of claims remain uncited or misattributed. This occurs because existing systems treat attribution as post-hoc annotation, thereby lacking structural guarantees.


\section{RegOps-Bench: Axis-Decoupled Construction for Compliance QA}
\label{sec:bench}

RegOps-Bench is designed around three core properties of this domain: (1) ground truth is a typed citation closure, (2) difficulty is defined by the structural complexity of the reference set rather than lexical phrasing, and (3) evaluation operates at the regulatory unit where authority delegates.

\begin{table}
\centering
\small
\caption{\textbf{RegOps-Bench corpus.} 12 Korean R\&D regulations spanning the statute-to-manual delegation chain. \#Art.\ counts articles; \#Prov.\ counts paragraphs, undefined (---) for the section-structured manual.}
\resizebox{\columnwidth}{!}{
\begin{tabular}{@{}llcrr@{}}
\toprule
Document & Authority Type & Tier & \#Art. & \#Prov. \\
\midrule
\multicolumn{5}{@{}l}{\textit{In-domain (7)}} \\
Innovation Act          & legal authority       & 1 & 42  & 161 \\
Enforcement Decree      & executive decree      & 2 & 75  & 225 \\
Enforcement Rules       & executive rule        & 3 & 4   & 7   \\
Cost-Use Standards      & admin. notice & 4 & 125 & 362 \\
Mgmt.\ Regulation       & admin. notice & 4 & 54  & 207 \\
Standard Guide          & admin. notice & 4 & 56  & 125 \\
Practitioner's Manual   & manual    & 5 & 40  & --- \\
\midrule
\multicolumn{5}{@{}l}{\textit{Auxiliary (2)}} \\
VAT Act                 & legal authority       & 1 & 85  & 294 \\
VAT Act Rules           & executive rule        & 3 & 85  & 153 \\
\midrule
\multicolumn{5}{@{}l}{\textit{Distractor (3)}} \\
S\&T Basic Act          & legal authority       & 1 & 71 & 208 \\
S\&T Basic Act Decree   & executive decree      & 2 & 72 & 209 \\
S\&T Basic Act Rules    & executive rule        & 3 & 9  & 16  \\
\midrule
Total & & & 718 & 1967 \\
\bottomrule
\end{tabular}
}
\label{tab:corpus}
\end{table}

\subsection{Operational Knowledge Graph (OKG)}
\label{sec:bench-corpus}

\urldef\iitpfaq\url{https://www.iitp.kr/web/lay1/bbs/S1T46C59/A/13/view.do?article_seq=4331&sort=latest&cpage=1&rows=10}

\paragraph{Corpus.} To anchor the benchmark in real-world compliance scenarios, we curated a corpus of 12 Korean R\&D regulatory documents covering the scope of 56 seed FAQs from an official FAQ document\footnote{\iitpfaq} of the Institute of Information \& Communications Technology Planning \& Evaluation (IITP). These documents span five authority tiers (1: statute -- 5: manual), from statutory acts down to operational manuals (Table~\ref{tab:corpus}). The seven in-domain documents form a largely self-contained delegation network. While certain high-level mandates inevitably delegate to external authorities, cross-document references, including all FAQ lineages, are mostly resolved within this set, where each lower tier procedurizes the open-textured mandates of the tier above it. A single compliance answer therefore requires composing a statutory obligation with its decree-level conditions and notice-level numeric thresholds. Two auxiliary VAT documents cover the tax-side cross-references recurring in cost-eligibility questions. The remaining three Science \& Technology Basic Act documents serve as distractors, sharing lexical overlap with the in-domain corpus but remaining legally out of scope.


\paragraph{OKG Construction.} 
We construct the OKG via deterministic rule extraction. By leveraging the formulaic citation patterns of legislation, this design effectively mitigates the generative hallucinations inherent in LLM-based indexing and provides deterministic edge typing—a prerequisite for evaluating closure. Furthermore, this approach eliminates heavy LLM computational overhead, making the index highly sustainable and easily updatable under frequent regulatory amendments. Each article (조) forms a node carrying its authority tier. Inter-article relations are typed into six classes: \texttt{PART\_OF} (hierarchy), \texttt{REFERENCES} (citation), the \texttt{DELEGATES\_TO}/\texttt{SPECIFIES} pair (downward delegation/upward realization), \texttt{DEFINES} (term-to-article), and \texttt{REQUIRES\_FORM} (article-to-form). For instance, ``as prescribed by the Presidential Decree'' yields a \texttt{DELEGATES\_TO} edge. Retrieval and end-to-end citation performance are primarily evaluated at the article level (조), the atomic unit of authority. Paragraph/item structure is retained within each node for condition evaluation, and Appendix~\ref{app:cit} additionally reports exact-granularity citation performance.

\subsection{QA Construction with Axis Decoupling}
\label{sec:bench-construction}

\begin{table}
\centering
\caption{\textbf{Difficulty rubric for RegOps-Bench.} Levels are checked from L4 to L1, and each question is assigned the highest difficulty level whose triggers it satisfies.}
\resizebox{\columnwidth}{!}{
\begin{tabular}{@{}lll@{}}
\toprule
Level & Characteristic & Trigger \\
\midrule
L1 & single-anchor lookup & $|\text{refs}|=1$, not conditional \\ \midrule
L2 & conditional or 2-ref & $|\text{refs}|\geq 2$ \textbf{or} conditional \\ \midrule
L3 & multi-hop / multi-doc & \makecell[l]{cross-doc, external-law,\\$|\text{refs}|\geq 4$, 4-institution parallel,\\or multi-facet arity $\geq 3$} \\ \midrule
L4 & conditional multi-hop & \makecell[l]{conditional \textbf{and} (cross-doc,\\sanction, multi-institution,\\or multi-facet arity $\geq 3$)} \\
\bottomrule
\end{tabular}
}
\label{tab:difficulty}
\end{table}

\paragraph{Axis-Decoupled Augmentation via LLM.} To ensure the benchmark comprehensively evaluates the structural complexity defined in our design principles, we expand the 56 seed FAQs into a final set of 231 expert-grounded questions using a high-capacity LLM (Gemini-3-flash~\cite{google2025gemini3flash}). Our augmentation follows an axis-decoupling principle with two distinct axes:
\begin{itemize}[leftmargin=*]\small
    \item \textbf{Question Type}\,---\,The substantive intent of the query (e.g., single-clause lookup, exception-heavy, or sanction-bearing condition).
    \item \textbf{Difficulty Level}\,---\,The structural complexity of the required reference set (L1 to L4), strictly bounded by the quantitative triggers in Table~\ref{tab:difficulty}.
\end{itemize}
The conditions are sampled and combined independently, so that a single question type can realize any of L1--L4 depending only on the anchor and injected facets, minimizing the confounding effect of surface-level phrasing.

For each seed, the LLM is given the governing articles and instructed to re-synthesize a situational, first-person practitioner inquiry matching the sampled axes, injecting facets (actors, temporal constraints, and institutional variables) that necessitate the target difficulty. For instance, to elevate a straightforward seed to L4, the model constructs a scenario in which an actor's specific condition triggers a cross-document citation to an external disqualification provision. Difficulty thus remains an intrinsic property of the regulatory logic (the reference set) rather than an artifact of linguistic complexity. To verify that our augmentation induces the targeted structural depth, we compare the corpus characteristics of the LLM-generated splits against the original human FAQs in Appendix~\ref{app:aug_valid}.


\paragraph{Reference Rules and Closure.} To reduce human annotator variance and ensure reproducibility, we standardize the ground-truth reference mapping via four deterministic expert rules, activated by formal structural markers identified during synthesis:
\begin{itemize}[leftmargin=*]\small
    \item \textbf{R1 (Domain Anchor):} binds the query to the top-priority controlling document of its domain---the Cost-Use Standards by default, or the domain-specific instrument otherwise (the Mgmt.\ Regulation for institutional-IT queries, the Standard Guide for facility/equipment, the VAT Act for tax-side, the Innovation Act for statutory-procedure).
    \item \textbf{R2 (Parallel \& Deemed-Application Expansion):} expands the set across the four institutional slots (government-funded, university, non-profit, for-profit) via parallel groups, and follows \emph{deemed-application} (의제 준용) forward edges so that a deemed article additionally pulls in the provisions it incorporates by reference.
    \item \textbf{R3 (Pre-Approval Exception):} binds exception-heavy inquiries to their governing pre-approval clause (Cost-Use Standards Art.~73).
    \item \textbf{R4 (Sanction Exhaustion):} closes sanction-bearing questions onto their corresponding settlement and disqualification provisions (Enforcement Decree Art.~26 and Cost-Use Standards Art.~83).
\end{itemize}

The external-law case is handled by absorbing it into the in-corpus VAT domain rather than as a separate rule, since the VAT Act and its Rules are already part of the corpus. This rule-based mapping turns each QA pair from an ambiguous retrieval task into a verifiable, deterministic traversal of the regulatory hierarchy.

\begin{figure*}
    \centering
    \includegraphics[width=1.\textwidth]{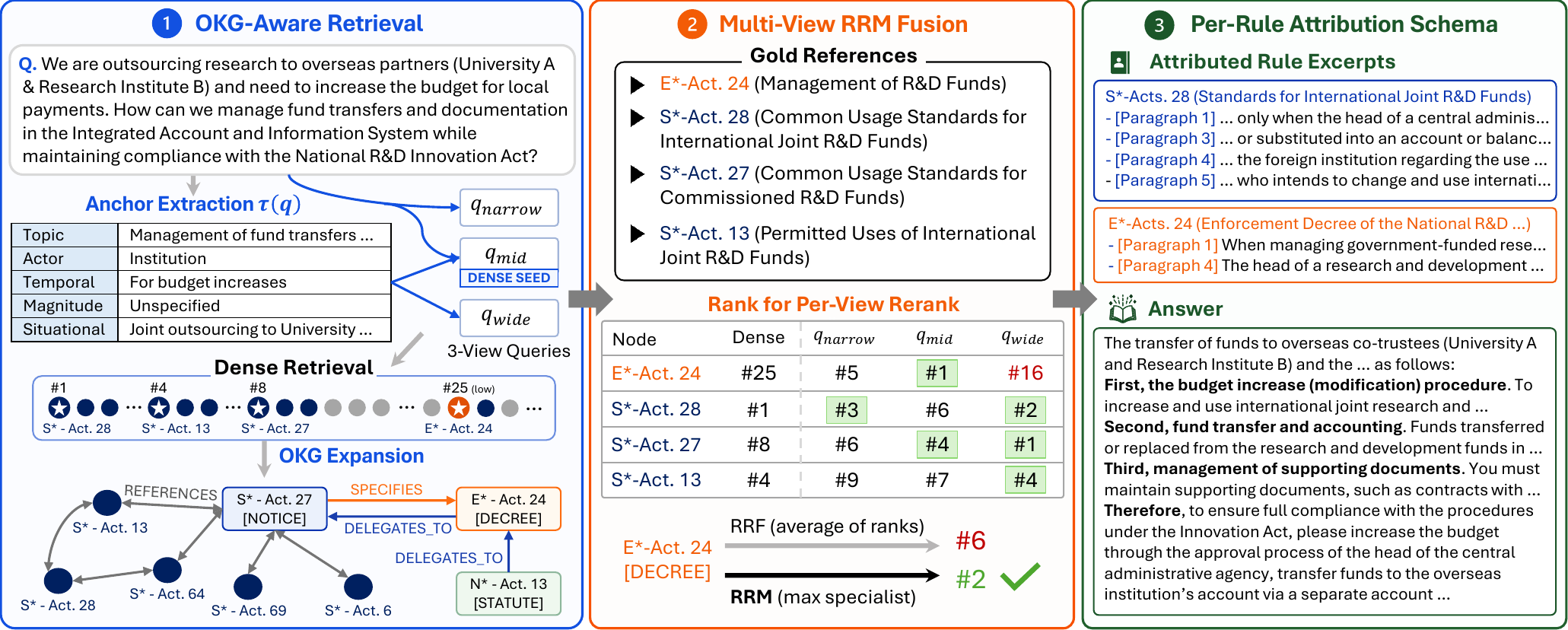}
    \caption{\textbf{Overall framework of RefWalk.} The examples are translated from the original Korean text for illustrative purposes. Document abbreviations are defined in Table 1 (\textbf{S*}: Standards for the Use of Expenses, \textbf{E*}: Enforcement Decree of Innovation Act, \textbf{N*}: Innovation Act).}
    \label{fig:pipeline}
\end{figure*}

\paragraph{Quality Assurance Protocol.} Each generated QA pair passes a layered validation cascade before merging. A regex runner gate first rejects forbidden formats, such as multi-institution enumerations and textbook-style summary or comparison requests. Survivors are scored by an LLM judge against a four-criterion rubric evaluating practitioner voice, concrete context, situational framing, and a single-institution viewpoint, yielding around a 50\% acceptance rate. This is supplemented by heuristic checks (script leakage, meta-patterns, degenerate length) and a difficulty re-validation step that discards items whose realized reference set diverges from the target level. A closure validator then verifies and auto-completes any missing R3 or R4-triggered authorities. We also conduct a manual curation pass over the accepted pool to filter residual low-quality items. Finally, ground-truth reference sets undergo a deterministic post-merge audit that corrects a clause-attribution artifact (cross-article paragraph-token leakage) present in 38\% of pre-audit items. The ground-truth reference sets are detailed in the Appendix~\ref{app:gt-refs}.


\section{RefWalk: Traversing Reference Paths for Structural Attribution}
\label{sec:method}

The rigorous properties of regulatory compliance QA—typed citation closures and article-level structural traversal—demand a framework where retrieval and generation operate as a traceable pipeline. RefWalk organizes this framework around the topic anchor, a structured abstraction distilling the query into its core procedural intent and conditional facets. Rather than treating query processing, graph expansion, and answer generation as isolated, fragmented stages, this single anchor propagates through the entire framework (Figure~\ref{fig:pipeline}). It simultaneously drives multi-view query construction, seeds OKG traversal, and re-enters the generation prompt to structurally constrain attribution.

\subsection{Topic-Anchored Multi-View Retrieval}
\label{sec:method-retrieval}

\paragraph{Anchor Extraction and Views.} For a given query $q$, a frozen LLM $\tau(q)$ extracts the topic anchor: a structured tuple of the core topic and facet conditions (actor, temporal, magnitude, and situational context). Because regulatory QA requires matching both explicit granular constraints and implicit procedural exceptions, we construct three targeted views from this shared anchor. $q_{\text{narrow}}$ directly uses the original question for dense semantic matching of explicit details. Conversely, $q_{\text{wide}}$ drops the raw question, relying solely on the abstracted topic and conditions to capture structural multi-hop references where surface phrasing is uninformative. $q_{\text{mid}}$ combines both for balanced retrieval.

\paragraph{OKG Expansion and RRM Fusion.} Retrieval begins by fetching an initial candidate pool $\mathcal{S}$ of size $N$ using $q_{\text{mid}}$ via dense retrieval. We expand this pool one hop along citation-bearing edges (\texttt{REFERENCES}, \texttt{DELEGATES\_TO}, \texttt{SPECIFIES}) in the OKG, isolating explicit delegations. To penalize indirect evidence, 1-hop neighbors receive a decay factor $\delta$. For nodes discovered via multiple pathways, we take the maximum decayed score to preserve the strongest signal without inflating it via sum-based aggregation. The expanded pool is then scored by a cross-encoder across our three semantic views and fused via Reciprocal Rank MAX (RRM):
\begin{equation}
\text{score}(d) \;=\; \max_{v \in \{narrow,\, mid,\, wide\}} \frac{1}{k + \text{rank}_v(d)},
\end{equation}
where $k$ denotes a smoothing constant. Unlike sum-based aggregations (e.g., RRF) that demand consensus, RRM ensures that candidates highly ranked by a single specialist view retain their priority. Finally, we introduce an authority-aware decay $\mu$ for candidates sourced from lower-tier operational manuals. Applied post-fusion, $\mu$ injects a domain-specific inductive bias into the ranking: it intrinsically prioritizes primary statutes, yet permits strongly matched manual passages to surface when they contain critical procedural details. This allows RefWalk to maintain citation closure while navigating multi-tiered delegations, circumventing the fragmentation of standard query decomposition.

\subsection{Per-Rule Attribution Schema}
\label{sec:method-attribution}

In regulatory compliance, the risk of a hallucinated citation outweighs the benefit of a broad response. Generative models typically treat attribution as a post-hoc afterthought, appending free-form footers that frequently mismatch the generated claims. RefWalk structurally mitigates this vulnerability through a per-rule attribution schema. The extracted topic anchor re-enters the framework by being injected into the generation prompt alongside the retrieved OKG passages and a strict JSON schema that maps specific \texttt{rule\_id} keys to arrays of generated claims. Rather than generating free-form text and retroactively appending citations, the model must emit procedural claims exclusively as array values bound to their governing rules. Under this strict schema, every generated claim is inherently bound to its source. This ensures that the structural precision achieved during OKG retrieval directly translates into the rigorous traceability required for regulatory compliance.

\section{Experiments}
\label{sec:experiments}

\subsection{Setup}
\label{sec:exp-setup}

\paragraph{Dataset \& Models.} To demonstrate the effect of our framework, we conduct our primary evaluations on RegOps-Bench. We also evaluate on a HIPAA-derived QA dataset ($n=100$) constructed following the De Jure procedure~\cite{guliani2026jure}. For retrieval, we use Qwen3-Embedding-0.6B and Qwen3-Reranker-0.6B~\cite{zhang2025qwen3} to ensure our gains stem from our structural design rather than brute-force embedding scale. For generations, we employ Qwen3.5-4B~\cite{qwen35blog} and Qwen3.6-35B~\cite{qwen36_35b_a3b} to observe scale-dependent behaviors, alongside Gemini-3.1-Pro~\cite{gemini31pro_modelcard} for frontier model validation.

\paragraph{Metrics.} For retrieval, we report Recall@10, nDCG@10, and FullCov@10. We introduce FullCov@10 (Full Coverage) as the fraction of queries where the entire ground-truth reference set is successfully retrieved within the top 10 candidates, effectively capturing the exhaustiveness required for compliance QA. End-to-end generation is evaluated using Claim F1 and Citation F1. Claim F1 applies an LLM judge~\cite{zheng2023judging} to label each predicted claim against the reference set as \{match, partial, none\}, with partials weighted 0.5 before bipartite resolution~\cite{min-etal-2023-factscore}. Citation F1 compares predicted and ground-truth references after rolling sub-article ids (항/호) up to their 조-level ancestor, so credit is given for citing the correct provision regardless of granularity~\cite{gao-etal-2023-enabling}. In regulatory compliance, Citation Precision is prioritized. While a missed citation merely results in an incomplete answer, hallucinating a legal citation poses severe operational risks. Thus, our analysis focuses on a model's ability to mitigate attribution hallucinations.

\begin{table}
\centering
\caption{\textbf{Retrieval performance on RegOps-Bench and HIPAA-derived QA (Top-10).}}
\label{tab:retrieval}
\resizebox{.91\columnwidth}{!}{
\begin{tabular}{lcccc}
\toprule
\multirow{2}{*}{Method} & \multicolumn{3}{c}{\textbf{RegOps}} & HIPAA \\ \cmidrule{2-4}
 & R & FullCov & nDCG & R\\
\midrule
LightRAG       & 45.3 & 32.5 & 31.9 & \textbf{100.0} \\
HippoRAG-2     & 42.6 & 29.9 & 26.5 & 97.0 \\
Query Decomp   & 50.8 & 33.8 & 42.7 & 92.0 \\
PIKE-RAG       & 54.6 & 37.2 & 51.4 & 97.0 \\ \midrule
BM25           & 34.8 & 25.5 & 27.0 & 94.0 \\
Dense          & 55.9 & 38.5 & 45.8 & 90.0 \\
Dense+Rerank   & 57.9 & 39.0 & 53.8 & 94.0 \\
\text{ }\text{ }w/ OKG & 60.6 & 43.3 & 55.8 & 94.0 \\
\midrule
\rowcolor{gray!10}
Ours           & \textbf{63.8} & \textbf{45.9} & \textbf{57.6} & 99.0 \\
\text{ }\text{ }w/o multi-view & \underline{62.5} & \underline{44.5} & 56.3 & 98.0 \\
\text{ }\text{ }w/o anchor  & 61.6 & 43.7 & \underline{57.2} & \textbf{100.0} \\
\bottomrule
\end{tabular}
}
\end{table}

\paragraph{Baselines.} We compare RefWalk against strong baselines across two stages. For retrieval, we evaluate standard architectures (BM25, Dense, Dense+Rerank) alongside state-of-the-art graph and query-based retrievers, including LightRAG \cite{guo-etal-2025-lightrag}, HippoRAG-2 \cite{gutierrez2025hipporag2}, Query Decomposition \cite{petcu-etal-2026-query}, and PIKE-RAG \cite{wang2025pike}. For end-to-end RAG, we evaluate NativeRAG with a Dense+Rerank pipeline as well as the four aforementioned systems equipped with generation capabilities. To ensure a fair comparison, all baselines share the exact same embedding, reranking, and generation backbones as RefWalk.

\paragraph{Hyperparameters.} For RefWalk, we set the retrieval pool size to $N=50$, the OKG seed count to $M=10$, and the RRM fusion constant to $k=60$. Further details are represented in Appendix~\ref{app:imp}.

\begin{table}[]
\caption{\textbf{End-to-end RAG performance on RegOps-Bench.} \textit{w/o} schema removes the per-rule attribution schema, emitting free-form output.}
\label{tab:rag}
\resizebox{1.\columnwidth}{!}{
\begin{tabular}{llcccccc}
\toprule
\multirow{2}{*}{Model} & \multirow{2}{*}{Method} & \multicolumn{3}{c}{Claim} & \multicolumn{3}{c}{Citation} \\ \cmidrule{3-5} \cmidrule{6-8}
                       &                         & F1      & P      & R      & F1       & P       & R       \\ \midrule
\multirow{6}{*}{\makecell{Qwen3.5\\4B}}      & NativeRAG               & 34.0    & 29.5   & \textbf{51.0}   & 36.6     & 37.7    & 45.9    \\
                       & LightRAG                & 30.4    & 25.8   & \underline{50.2}   & 27.1     & 24.2    & 42.1    \\
                       & HippoRAG-2              & 34.8    & 31.4   & 49.5   & 36.8     & 37.7    & \underline{46.7}    \\
                       & PIKE-RAG                & \underline{35.6}    & \underline{38.7}   & 40.0   & \underline{45.0}     & \underline{54.2}    & 45.9    \\ \cmidrule{2-8}
                       & Ours  & \textbf{36.4}    & \textbf{40.9}   & 39.6   & \textbf{47.0}     & \textbf{57.1}    & 46.5  \\
                       & w/o schema     & 34.7    & 31.3   & 49.4   & 40.2     & 40.0    & \textbf{51.1}    \\ \midrule
\multirow{7}{*}{\makecell{Qwen3.6\\35B}}            & NativeRAG               & 37.7    & 35.8   & 51.2   & 46.7     & 52.3    & 50.6    \\
                       & LightRAG                & 37.0    & 33.6   & \textbf{53.1}   & 40.7     & 39.0    & 51.9    \\
                       & HippoRAG-2              & 37.0    & 34.8   & 50.4   & 47.8     & 53.1    & 51.9    \\
                       & PIKE-RAG                & \underline{38.0}    & \underline{42.6}   & 41.2   & 44.3     & 56.2    & 45.3    \\ \cmidrule{2-8}
                       & Ours                    & \textbf{41.1}    & \textbf{43.2}   & 47.4   & \textbf{54.9}     & \textbf{68.0}    & \underline{53.6}    \\
                       & w/o schema              & 37.9    & 36.1   & \underline{51.3}   & \underline{51.8}     & \underline{57.0}    & \textbf{56.3}    \\
\bottomrule                       
\end{tabular}
}
\vspace{-3pt}
\end{table}


\subsection{Retrieval Results}
\label{sec:retrieval-results}

Table~\ref{tab:retrieval} presents the retrieval performance. RefWalk achieves the best performance among the evaluated methods on RegOps-Bench. While these absolute metrics leave an unsolved gap for future work, this steep difficulty highlights the necessity of RegOps-Bench as a non-saturated stress test rather than a limitation of our method. Ablation studies confirm that removing multi-view reranking, OKG expansion, or anchor enrichment each degrades performance, demonstrating that these components jointly address the structural failures of dense retrieval. On the HIPAA dataset, retrieval metrics generally saturate across most baselines. This suggests that while baseline retrieval is highly effective on flat-structure benchmarks, simple semantic matching is insufficient for the complex cross-reference environments.

\begin{table}[]
\centering
\caption{\textbf{End-to-end RAG performance on HIPAA-derived QA (cross-domain).} }
\label{tab:HIPAA}
\resizebox{.92\columnwidth}{!}{
\begin{tabular}{llcc}
\toprule
Model                        & Method     & Claim F1 & Citation F1 \\ \midrule
\multirow{6}{*}{Qwen3.5-4B}  & NativeRAG   & 62.3     & 72.8        \\
                             & LightRAG    & 47.4     & 49.8        \\
                             & HippoRAG-2  & 58.8     & 68.6        \\
                             & PIKE-RAG    & \textbf{73.9}     & \underline{80.6}        \\ \cmidrule{2-4}
                             & Ours        & \underline{72.4}     & \textbf{84.0}        \\ 
                             & w/o schema     & 58.3     & 70.6        \\ \midrule
\multirow{6}{*}{Qwen3.6-35B} & NativeRAG   & 61.8     & 73.7        \\
                             & LightRAG    & 50.0     & 48.1        \\
                             & HippoRAG-2  & 62.0     & 74.0        \\
                             & PIKE-RAG    & \underline{71.9}     & \underline{82.2}        \\ \cmidrule{2-4}
                             & Ours       & \textbf{73.9} & \textbf{85.3}        \\ 
                             & w/o schema    & 66.4 & 73.3      \\ 
                             \bottomrule
\end{tabular}
}
\end{table}

\subsection{End-to-End RAG Results}
\label{sec:rag-results}

Table~\ref{tab:rag} compares the overall end-to-end performance on RegOps-Bench. While Claim F1 remains relatively consistent across systems—since powerful generators can infer similar claims once partial evidence is retrieved—RefWalk achieves substantial gains in Citation F1 over Native RAG and graph-based baselines. This improvement is structurally enforced by our per-rule attribution schema. We validate this through the \texttt{w/o schema} ablation, which effectively isolates the generation bottleneck by pairing RefWalk's advanced retrieval with a Native RAG-style free-form prompt. When the schema constraint is removed, Citation F1 drops sharply. This result shows that without schema-level structural binding, generative models treat citations as post-hoc footers, failing to accurately map claims to their corresponding rules even when the correct evidence is retrieved.

\subsection{Difficulty-Stratified Analysis}
\label{sec:exp-difficulty}

\begin{table}[t]
\centering
\footnotesize
\setlength{\tabcolsep}{5pt}
\caption{Retrieval performance by query difficulty (L1--L4) on the
RegOps benchmark. We report R@10 and FullCov@10 (\%). Best per column is
in \textbf{bold}, second-best is \underline{underlined}.}
\label{tab:retrieval_per_difficulty}
\resizebox{\columnwidth}{!}{%
\begin{tabular}{l rrrr rrrr}
\toprule
& \multicolumn{4}{c}{\textbf{R@10}} & \multicolumn{4}{c}{\textbf{FullCov@10}} \\
\cmidrule(lr){2-5} \cmidrule(lr){6-9}
\textbf{Method} & L1 & L2 & L3 & L4 & L1 & L2 & L3 & L4 \\
\midrule
Dense        & \underline{86.5} & 64.0 & 35.2 & 35.0 & \underline{85.4} & 48.1 & \underline{13.3} & 5.3 \\
LightRAG     & 79.2 & 52.9 & 24.8 & 22.1 & 77.1 & 38.3 & \underline{13.3} & 1.8 \\
HippoRAG-2   & 77.1 & 49.6 & 22.2 & 19.9 & 75.0 & 32.1 & \underline{13.3} & 1.8 \\
Query Decomp & 83.3 & 55.8 & 32.1 & 31.1 & 81.2 & 40.7 & 8.9 & 3.5 \\
PIKE-RAG     & 75.0 & \underline{64.6} & \underline{35.3} & \underline{38.6} & 75.0 & \underline{49.4} & 11.1 & \underline{8.8} \\ \midrule
\textbf{Ours} & \textbf{90.6} & \textbf{73.3} & \textbf{44.0} & \textbf{43.4} & \textbf{89.6} & \textbf{59.3} & \textbf{20.0} & \textbf{10.5} \\
\bottomrule
\end{tabular}%
}
\end{table}

\begin{table}
\centering
\footnotesize
\setlength{\tabcolsep}{5pt}
\caption{End-to-end RAG performance by query difficulty (L1--L4) on the RegOps benchmark
($n=48/81/45/57$ for L1/L2/L3/L4).}
\label{tab:e2e_rag_per_difficulty}
\resizebox{\columnwidth}{!}{%
\begin{tabular}{l rrrr rrrr}
\toprule
& \multicolumn{4}{c}{\textbf{Claim F1}} & \multicolumn{4}{c}{\textbf{Citation F1}} \\
\cmidrule(lr){2-5} \cmidrule(lr){6-9}
\textbf{Method} & L1 & L2 & L3 & L4 & L1 & L2 & L3 & L4 \\
\midrule
\multicolumn{9}{l}{\textit{Backbone: Qwen3.5-4B}} \\
NativeRAG            & 40.6 & 40.1 & 24.0 & 27.8 & 54.4 & 39.7 & 24.3 & 26.9 \\
LightRAG             & 30.2 & 35.9 & 23.8 & \underline{28.0} & 38.1 & 29.2 & 18.5 & 21.8 \\
HippoRAG-2           & 42.3 & 39.1 & \underline{25.1} & \textbf{30.0} & 57.8 & 40.1 & 21.7 & 26.5 \\
PIKE-RAG             & \textbf{45.4} & \underline{43.2} & 21.9 & 27.5 & \underline{67.0} & \underline{55.4} & \underline{24.5} & \underline{27.7} \\ \midrule
Ours & \underline{43.1} & \textbf{46.0} & \textbf{26.1} & 25.3 & \textbf{68.4} & \textbf{57.8} & \textbf{27.0} & \textbf{29.2} \\
\midrule
\multicolumn{9}{l}{\textit{Backbone: Qwen3.6-35B}} \\
NativeRAG            & 44.7 & 42.1 & \underline{28.6} & \underline{32.7} & 68.2 & \underline{55.2} & 29.3 & 30.1 \\
LightRAG             & 42.3 & 44.5 & 28.0 & 29.0 & 58.1 & 50.9 & 23.9 & 25.0 \\
HippoRAG-2           & 48.5 & 41.8 & 28.2 & 27.4 & \underline{75.0} & 53.7 & \underline{29.7} & 31.0 \\
PIKE-RAG             & \underline{49.1} & \underline{45.5} & 26.2 & 27.5 & 62.3 & 52.9 & 25.4 & \underline{31.9} \\ \midrule
Ours & \textbf{49.3} & \textbf{48.5} & \textbf{29.1} & \textbf{33.0} & \textbf{79.0} & \textbf{66.7} & \textbf{33.6} & \textbf{34.7} \\
\bottomrule
\end{tabular}%
}
\end{table}

To understand exactly where existing systems fail, we stratify performance by difficulty (L1–L4). As shown in Table \ref{tab:retrieval_per_difficulty}, all methods succeed at single-anchor lookups (L1). However, at L3 (multi-hop/cross-doc) and L4 (conditional multi-hop), baseline retrieval drops sharply. Lexical or simple dense retrieval cannot navigate explicit, multi-tiered delegations. This bottleneck directly cascades into generation, as demonstrated in Table \ref{tab:e2e_rag_per_difficulty}. While baseline models attempt to answer L3/L4 queries, their Citation F1 plummets because they fail to bind claims to the correct cross-document sources. In contrast, RefWalk demonstrates consistent improvements in Citation F1 across these complex tiers, effectively mitigating the severe drop-offs caused by fragmented retrieval.

 \begin{table}[]
\centering
\caption{\textbf{Mechanism Analysis of Multi-View Fusion.} RRM maximizes overall performance by acting as a specialist selector.}
\label{tab:rrm-vs-rrf}
\resizebox{\columnwidth}{!}{
\begin{tabular}{lccccc}
\toprule
\multirow{2}{*}{Variant} & Overall & \multicolumn{4}{c}{Recall@10 by Difficulty} \\ \cmidrule{3-6}
 & R@10 & L1 & L2 & \textbf{L3} & L4 \\
\midrule
\rowcolor{gray!10}
3-view RRM (Ours) & \textbf{63.8} & 90.6 & 73.3 & \textbf{44.0} & 43.4 \\
3-view wRRF (wide=2) & 62.5 & 87.5 & 73.7 & 39.5 & 43.6 \\
3-view RRF (uniform) & 62.9 & \textbf{91.7} & \textbf{74.1} & 36.2 & \textbf{43.9} \\
\midrule
wide-only & 61.9 & 88.5 & 73.3 & 38.9 & 41.3 \\
narrow-only & 61.7 & 90.6 & 71.0 & 37.4 & 43.4 \\
\bottomrule
\end{tabular}
}
\end{table}

\subsection{Mechanism Analysis: RRM vs RRF}
\label{sec:exp-mechanism}

Table~\ref{tab:rrm-vs-rrf} suggests that RRM’s overall advantage is primarily attributable to L3 queries. This pattern is consistent with max-based fusion preserving a strong specialist signal when relevance is asymmetric across views, whereas sum-based fusion may dilute that signal through cross-view aggregation. Its comparable performance at the other tiers further indicates that the benefit is associated with view disagreement rather than query difficulty alone.

\subsection{Mitigating Attribution Hallucination via Schema Constraints}

To validate whether per-rule schema alleviates attribution hallucination (Table \ref{tab:generation_reliability}), we conduct a ceiling analysis using an Oracle with Controlled Noise setting ($k=10$). While Native RAG's scores are artificially inflated under a Pure Oracle setting (without distractors), padding the context with hard negatives from the 1-hop citation neighborhood exposes a systemic eager-citing bias, causing a sharp decline in Citation Precision. Conversely, RefWalk's conservative mapping strategy filters out these hard negatives, securing a highly competitive Citation Precision. Applying our schema constraint to PIKE-RAG also boosts its native Claim Precision and F1. This highlights the transferability of our schema-binding strategy in steering LLMs toward more precise operational claims. However, even with this structural enforcement, PIKE-RAG's Citation F1 remains substantially lower than RefWalk's. Furthermore, evaluating Gemini-3.1-Pro reveals that even frontier models with advanced reasoning capabilities are highly susceptible to attribution hallucination under Native RAG.

\begin{table}
\caption{\textbf{Effect of Schema Constraint on Attribution Reliability.} We compare generation methods across varying model capacities and retrieval settings.}
\label{tab:generation_reliability}
\resizebox{1.\columnwidth}{!}{
\begin{tabular}{lllcccc}
\toprule
\multirow{2}{*}{Model} & \multirow{2}{*}{Retrieval} & \multirow{2}{*}{Method} & \multicolumn{2}{c}{Claim} & \multicolumn{2}{c}{Citation} \\ \cmidrule(lr){4-5} \cmidrule(lr){6-7}
                       &                            &                         & P      & F1     & P      & F1      \\ \midrule
\multirow{9}{*}{\makecell{Qwen3.6\\35B}} 
    & \multirow{2}{*}{Top-10}    
        & Native               & 35.8   & 37.7   & 52.3   & 46.7    \\
    &   & Ours & \textbf{43.2} & \textbf{41.1} & \textbf{68.0} & \textbf{54.9} \\ \cmidrule{2-7}
    & \multirow{2}{*}{Pure Oracle} 
        & Native               & 47.5   & 44.0   & \textbf{100.0}* & \textbf{82.9}* \\
    &   & Ours & \textbf{50.0} & \textbf{44.9} & 99.6*   & 76.0*    \\ \cmidrule{2-7}
    & \multirow{2}{*}{\makecell{Oracle\\($k=10$)}} 
        & Native               & 40.7   & 40.9   & 82.3   & \textbf{70.3} \\
    &   & Ours & \textbf{47.2} & \textbf{43.1} & \textbf{85.2} & 67.4    \\ \cmidrule{2-7}
    & \multirow{2}{*}{Top-10} & PIKE-RAG & \multirow{2}{*}{45.0} & \multirow{2}{*}{41.0} & \multirow{2}{*}{64.3} & \multirow{2}{*}{47.5} \\
    &  & w/ schema &  &  &  &  \\ \midrule
\multirow{2}{*}{\makecell{Gemini\\3.1-Pro}}  
    & \multirow{2}{*}{Top-10}    
        & Native               & 48.3   & \textbf{41.7}   & 52.6   & 46.2    \\
    &   & Ours & \textbf{53.7} & \textbf{41.7} & \textbf{65.7} & \textbf{54.6} \\ \bottomrule
\multicolumn{7}{l}{\small *Artificially inflated due to the absence of distractors.}
\end{tabular}
}
\end{table}

\section{Conclusion}

Assisting regulatory compliance with LLMs requires a paradigm shift from broad answer generation to verifiable, structure-bound attribution. In this work, we formalized this challenge through RegOps-Bench, a benchmark designed to capture the intricate, multi-tiered delegation networks inherent in real-world regulatory ecosystems. By decoupling procedural complexity from surface-level phrasing, our evaluations revealed that current retrieval-augmented systems—despite their proficiency on flat-structure tasks—struggle significantly to navigate complex legal citations. To overcome this, we introduced RefWalk, a framework that unifies Operational Knowledge Graph traversal with per-rule attribution. Our findings demonstrate that preserving specialist signals via multi-view RRM fusion improves the recovery of cross-document chains, while structural schema-binding curtails the persistent attribution hallucinations prevalent even in frontier LLMs. Ultimately, RefWalk successfully mitigates the vulnerabilities of free-form generation, providing a robust and generalizable foundation for advancing more traceable AI in high-stakes regulatory domains.

\section{Limitations}

While RefWalk establishes a robust framework for verifiable compliance QA, its design philosophy fundamentally prioritizes safety over comprehensiveness, introducing an inherent precision-recall trade-off in attribution. In high-stakes domains like regulatory compliance, hallucinating a false citation is far more dangerous than missing a valid one. Consequently, our per-rule attribution schema enforces a highly conservative mapping strategy. As demonstrated in our oracle experiments, this strictness successfully pushes citation precision to highly reliable levels under noisy conditions, but it inevitably sacrifices recall compared to the eager-citing behavior of Native RAG. Future work should explore adaptive schema constraints that dynamically balance recall without compromising strict hallucination boundaries.

Beyond generation constraints, the framework's retrieval architecture relies on the deterministic extraction of an Operational Knowledge Graph (OKG). By leveraging the highly formulaic citation patterns of Korean regulatory documents, we achieved high-precision, low-cost rule extraction without the heavy LLM-based indexing overhead seen in other graph-based RAGs. However, scaling this purely deterministic graph-building process to less structured jurisdictions, such as heavily case-law-driven domains or entirely different languages, remains a challenge. Adapting RefWalk to such environments will likely require transitioning to hybrid (rule and LLM-assisted) extraction pipelines.

Finally, navigating the OKG accurately requires preserving expert signals during complex multi-hop retrieval, which RefWalk achieves via multi-view cross-encoding and RRM fusion. While highly effective for targeted candidate pools (e.g., $N=50$), applying deep cross-attention independently across three distinct semantic views scales linearly with the pool size, which may eventually encounter computational bottlenecks in massive-scale deployments. Addressing these latency challenges—such as optimizing prompt topology for cross-view KV cache sharing, or introducing dynamic view routing to conditionally bypass redundant cross-encoder passes—remains an essential direction for scaling RefWalk without sacrificing its rigorous matching precision.

\section{Acknowledgments}

This work was supported by the Institute of Information \& Communications Technology Planning \& Evaluation (IITP) grants funded by the Korea government (MSIT) (No. RS-2019-II190079 (Artificial Intelligence Graduate School Program (Korea University)), No. IITP-2025-RS-2024-00436857 (Information Technology Research Center (ITRC)), No. IITP-2026-RS-2025-02304828 (Artificial Intelligence Star Fellowship Support Program to Nurture the Best Talents), and No. RS-2026-25522672 (Development of Unified Reasoning Technology Mimicking Human Cognition for Hierarchical Understanding and Unbounded Problem Solving)).

\bibliography{custom}

\appendix

\section{Appendix}
\label{sec:appendix}

\subsection{Implementation Details}
\label{app:imp}

All open-source models were served using vLLM~\cite{kwon2023efficient} on RTX A6000 (48 GB) GPUs in bfloat16 precision with a maximum sequence length of 32,768 tokens. We employed \texttt{Qwen3.6-35B-A3B-FP8} and \texttt{Qwen3.5-4B} as the 35B and 4B backbones, respectively, for the cross-scale comparison. Dense retrieval was conducted using \texttt{Qwen3-Embedding-0.6B}, while the cross-encoder utilized \texttt{Qwen3-Reranker-0.6B} with the official yes/no template. For all text generation, the temperature, top-$p$, and top-$k$ parameters were set to 0.1, 0.95, and 20, respectively. We disabled the thinking mode and fixed the random seed to 42 across all runs. For the retrieval stage of RefWalk, we utilized a 50-candidate rerank pool over three views fused via Reciprocal Rank MAX (k=60). This was further augmented by a 1-hop OKG expansion over \texttt{REFERENCES}, \texttt{DELEGATES\_TO}, and \texttt{SPECIFIES} edge types, applying a decay factor of $\delta=0.7$ and up to $M=10$ seed nodes. Here, $\delta$ down-weights 1-hop neighbors as indirect evidence; co-discovered neighbors take the max decayed score, not the sum. In addition, an authority-aware manual-node decay of $\mu=0.7$ is multiplied into the fused rerank score of any candidate sourced from a manual (\textit{매뉴얼}) rather than a statute (\textit{법령/시행령/시행규칙}). Because it is applied after multi-view fusion, it acts as a soft tie-break that favors higher-authority sources while still allowing a strongly-matched manual passage to outrank a weakly-matched statute. All hyperparameters were kept fixed across both the RegOps and HIPAA datasets. End-to-end generation is evaluated using Claim F1 and Citation F1. Claim F1 is computed via LLM-as-judge~\cite{zheng2023judging} bipartite matching between predicted and reference atomic claims, where exact semantic matches receive full credit and partial matches—those preserving the core proposition but missing a condition, exception, or numeric scope—receive half credit. Citation F1 measures the overlap between cited and ground-truth references at the article (조) level, rolling up sub-article citations (항/호) to their parent article so that an answer is credited whenever it points to the correct legal provision regardless of granularity.

\subsection{Knowledge Source and OKG Properties}
\label{app:corpus-okg}

\paragraph{Statistics.} The underlying corpus of RegOps-Bench comprises 718 procedural articles containing roughly 478K subword tokens. As detailed in Table~\ref{tab:corpus-content}, the article length distribution is heavily right-skewed. However, because only a small fraction of the articles exceed the standard 8,192-token embedding context window, article-level indexing remains lossless for the overwhelming majority of the text. Furthermore, the corpus includes a non-trivial amount of structured data, with approximately 10\% of articles embedding tabular content and 3\% acting as appendix forms (별표).

\paragraph{OKG Composition.} The Operational Knowledge Graph (OKG) instantiated from this textual corpus consists of 2,572 nodes (primarily operational article/provision units) and 3,942 typed edges. As Table~\ref{tab:okg-edges} illustrates, the graph's edge distribution is dominated by the structural containment hierarchy (\texttt{Part\_Of}) and citation-bearing relations. Notably, the upward realization (\texttt{Specifies}) and downward delegation (\texttt{Delegates\_To}) edges are perfectly balanced. This symmetry reflects the paired multi-tier legal structure of the corpus, forming the exact structural signature that retrieval systems must traverse to resolve complex regulatory queries.

\paragraph{Reference Topology of the OKG.} Beyond local edge distributions, we analyze the global connectivity of the OKG to substantiate that the in-domain documents form a largely self-contained delegation network. Considering only the citation-bearing relations (excluding the within-document containment hierarchy), the graph fragments into 1,878 components but is dominated by a single referential backbone of 654 nodes spanning all twelve documents; the remaining components have a median size of one, indicating that most articles delegate into the shared backbone rather than isolated islands. When the containment hierarchy is added, the graph consolidates into 148 components. This includes a giant component of 2,078 nodes that again spans all twelve documents, leaving only 81 truly orphaned nodes. Of the 148 components, only nine bridge two or more documents, mostly representing auxiliary tax pairings. Because the giant component co-mingles the containment hierarchy with citation edges, we further apply greedy-modularity community detection to it, recovering 48 distinct communities. The largest communities align tightly with intuitive procedural themes: a statutory-procedure cluster anchored on the Innovation Act (314 nodes across 9 documents), a cost-use cluster anchored on the Cost-Use Standards (300 nodes, 6 documents), and a self-contained tax cluster anchored on the VAT Act (264 nodes, 1 document). This empirical decomposition confirms that the benchmark's multi-hop reference closures are not artifacts of a single dense hub but rather span semantically coherent, multi-document procedural neighborhoods.

\begin{table}[ht]
\centering
\caption{Per-article length distribution of the knowledge source. Tokens are computed with the Qwen3-Embedding tokenizer. Sub-items count the paragraph/subparagraph (항/호) units nested in each article.}
\resizebox{\columnwidth}{!}{
\begin{tabular}{lrrrrr}
\toprule
Metric & Mean & p50 & p75 & p95 & Max \\
\midrule
Characters / article      & 993.1 & 422 & 847 & 2567 & 25226 \\
Tokens / article          & 665.7 & 290 & 576 & 1753 & 16801 \\
Sub-items / article       & 2.7   & 2   & 4   & 8    & 17 \\
\bottomrule
\end{tabular}
}
\label{tab:corpus-content}
\end{table}

\begin{table}[ht]
\centering
\small
\caption{Typed-edge composition of the OKG. The balanced \textsc{Delegates\_To}/\textsc{Specifies} counts reflect the paired delegation/realization structure of the corpus.}
\begin{tabular}{lrr}
\toprule
Edge type & Count & Ratio (\%) \\
\midrule
\texttt{Part\_Of} (hierarchy)        & 1730 & 43.9 \\
\texttt{References} (citation)       & 881  & 22.3 \\
\texttt{Specifies} (upward)          & 557  & 14.1 \\
\texttt{Delegates\_To} (downward)    & 557  & 14.1 \\
\texttt{Requires\_Form}              & 142  & 3.6 \\
\texttt{Defines}                     & 75   & 1.9 \\
\midrule
Total                                & 3942 & 100.0 \\
\bottomrule
\end{tabular}
\label{tab:okg-edges}
\end{table}

\subsection{Benchmark Details}
\label{app:bench-composition-structure}

\begin{table}[ht]
\centering
\small
\caption{Composition of the RegOps-Bench questions. All percentages are over the full set; under ``Domain anchor'' the 56 human seeds (24.2\%) carry no synthesized anchor and are omitted from the listing.}
\resizebox{\columnwidth}{!}{
\begin{tabular}{llr r}
\toprule
Axis & Category & $n$ & \% \\
\midrule
\multirow{4}{*}{Difficulty}
  & L1 & 48 & 20.8 \\
  & L2 & 81 & 35.1 \\
  & L3 & 45 & 19.5 \\
  & L4 & 57 & 24.7 \\
\midrule
\multirow{6}{*}{Question type}
  & IITP seed (human)       & 56 & 24.2 \\
  & Multi-facet             & 53 & 22.9 \\
  & Single-clause           & 50 & 21.6 \\
  & General-principle       & 24 & 10.4 \\
  & Exception-paired        & 26 & 11.3 \\
  & Institution-specific    & 22 & 9.5 \\
\midrule
\multirow{2}{*}{Source}
  & Augmented (LLM)         & 175 & 75.8 \\
  & IITP board (human)      & 56  & 24.2 \\
\midrule
\multirow{5}{*}{Domain anchor}
  & Cost-Use Standards (default) & 103 & 44.6 \\
  & Innovation-Act procedure     & 26  & 11.3 \\
  & Institutional-IT             & 22  & 9.5 \\
  & Facility/equipment           & 15  & 6.5 \\
  & Tax (VAT)                    & 9   & 3.9 \\
\bottomrule
\end{tabular}
}
\label{tab:bench-comp}
\end{table}

RegOps-Bench comprises 231 question--answer pairs, combining 175 instances generated via axis-decoupled augmentation with 56 verbatim queries retained from the IITP practitioner board. As outlined in Table~\ref{tab:bench-comp}, the benchmark spans diverse substantive intents---ranging from single-clause lookups to multi-facet institutional inquiries. The Cost-Use Standards serves as the primary domain anchor for 44.6\% of the full benchmark, supplemented by procedural chains from the Innovation Act, institutional IT regulations, and auxiliary tax documents.

We balance the dataset toward higher complexity, with the advanced L3 and L4 tiers jointly accounting for 44.2\% of the benchmark. This distribution ensures the evaluation stresses multi-hop and conditional reasoning rather than simple fact retrieval. Table~\ref{tab:diff-structure} shows an overall increase in structural complexity across the tiers. Mean reference-set size grows from 1.31 at L1 to 6.11 at L4, and cross-document references become prevalent at L3 and L4. Conditionality is concentrated in L2 and becomes mandatory at L4 by design, while mean graph depth remains unchanged at 3.16 across L3 and L4.

\begin{figure*}[t]
    \centering
    \includegraphics[width=\textwidth]{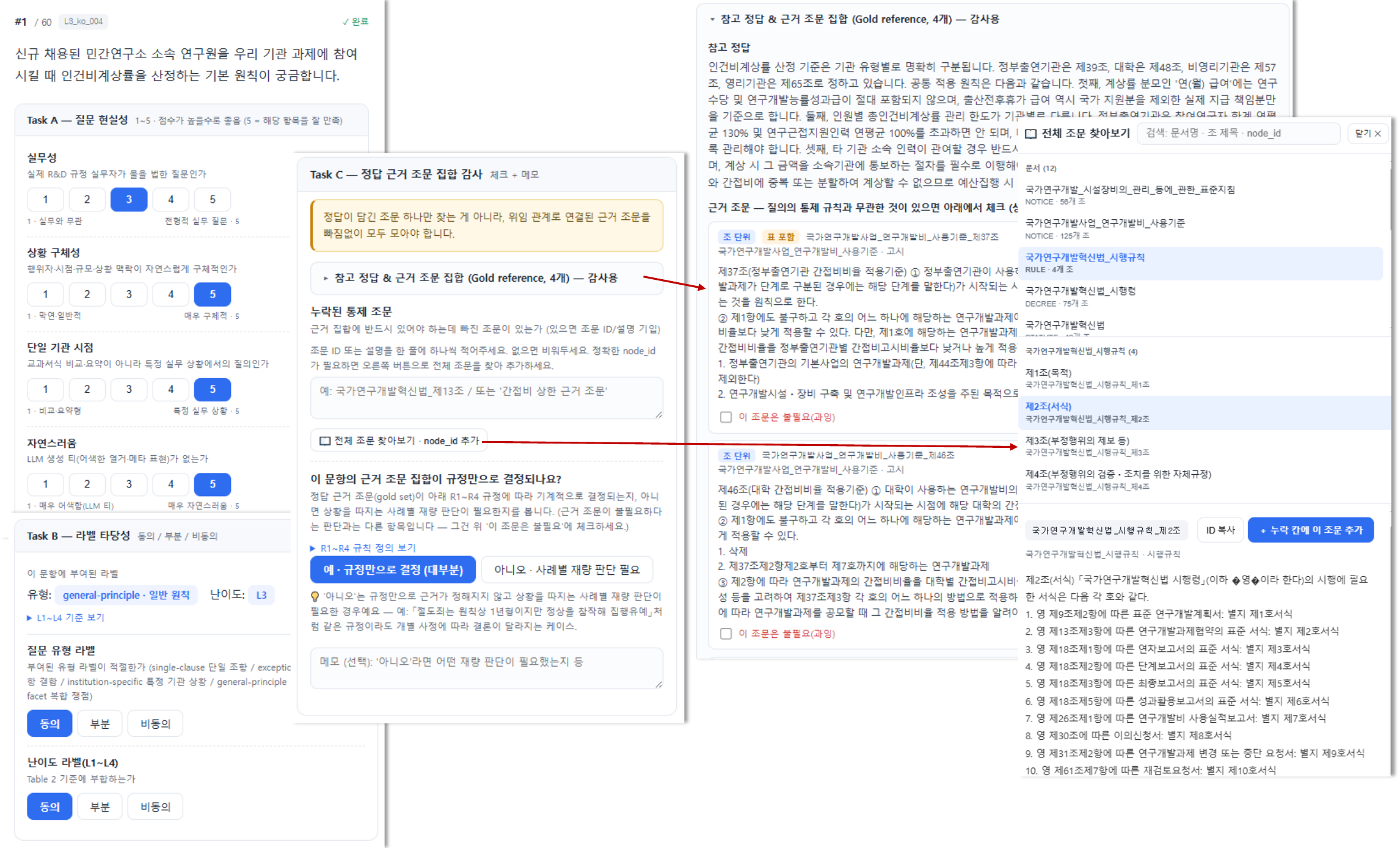}
    \caption{Korean-language annotation interface used for human validation of RegOps-Bench. Validators (A) rated question realism, situational specificity, single-institution framing, and naturalness; (B) assessed the validity of question-type and difficulty labels; and (C) audited the completeness of the gold reference set. The interface displayed the reference answer and current gold provisions and allowed validators to search the full corpus and add missing nodes. Red arrows highlight the controls for inspecting and adding provisions.}
    \label{fig:survey}
\end{figure*}

\paragraph{Human Validation.} We additionally conduct human validation to assess whether the augmented questions preserve the practical characteristics of real-world regulatory inquiries and whether their structural labels and reference sets are reliable. Prior to item-level validation, two domain experts with at least four years of experience in national R\&D administration reviewed the deterministic construction rules, including the facet axes, parallel-citation groups, and delegation triggers used to derive the reference sets. For item-level validation, four national R\&D practitioners independently evaluated a difficulty$\times$source-stratified 60-question subset, with every annotator reviewing every item. The evaluation set contains 47 augmented questions and 13 blinded controls taken verbatim from the official IITP FAQ. Annotators assessed question realism, the validity of the assigned question type and difficulty, and the completeness of the governing reference set. All validators were informed that their annotations would be used for research and reported only in aggregate form, and they consented to this use before participating. Figure~\ref{fig:survey} presents the annotation interface and task-specific instructions provided to the validators. As shown in Table~\ref{tab:human-validation}, augmented questions receive realism ratings comparable to the blinded official FAQs, indicating that the augmentation preserves the characteristics of practitioner inquiries. Annotators also show high agreement with the assigned question types and difficulty levels. For reference-set validation, most individual audits judge the governing rules sufficient, and no question receives a consensus missing-rule flag. The remaining disagreement is primarily due to over-inclusion, suggesting that residual annotation noise is concentrated on reference precision rather than missing governing evidence.

\begin{table}[ht]
\centering
\small
\caption{Structural profile of RegOps-Bench by difficulty. ``Cond.''\ = conditional, ``X-doc''\ = cross-document reference, ``Chain''\ = reference set forms a procedure chain; ``Refs''\ and ``Hops''\ are the mean realized reference-set size and mean citation-graph depth.}
\begin{tabular}{lrrrrr}
\toprule
& \multicolumn{1}{c}{Cond.} & \multicolumn{1}{c}{X-doc} & \multicolumn{1}{c}{Chain} & \multicolumn{1}{c}{Refs} & \multicolumn{1}{c}{Hops}\\
Level & \% & \% & \% & avg & avg \\
\midrule
L1 & 0.0   & 0.0  & 0.0  & 1.31 & 1.00 \\
L2 & 87.7  & 0.0  & 33.3 & 2.19 & 1.43 \\
L3 & 11.1  & 55.6 & 82.2 & 5.69 & 3.16 \\
L4 & 100.0 & 66.7 & 93.0 & 6.11 & 3.16 \\
\bottomrule
\end{tabular}
\label{tab:diff-structure}
\end{table}

\begin{table}[t]
\centering
\small
\setlength{\tabcolsep}{4.5pt}
\caption{
Human validation of RegOps-Bench. The evaluation uses a stratified subset containing 13 official FAQ questions and 47 augmented questions. Consensus reference errors require agreement from at least two annotators.
}
\label{tab:human-validation}
\resizebox{\columnwidth}{!}{
\begin{tabular}{lccc}
\toprule
\textbf{Metric} &
\textbf{Official} &
\textbf{Augmented} &
\textbf{Overall} \\
\midrule

\multicolumn{4}{l}{\textit{Question realism}} \\
Mean rating ($1$--$5$)       & 4.71 & 4.67 & 4.68 \\
Items with mean $\geq 4$     & 100.0 & 100.0 & 100.0 \\

\midrule
\multicolumn{4}{l}{\textit{Structural-label validity}} \\
Question-type agreement      & -- & 93.1 & -- \\
Difficulty-label agreement   & -- & 89.4 & -- \\
Pairwise agreement (type)    & -- & 86.2 & -- \\
Pairwise agreement (difficulty) & -- & 80.1 & -- \\

\midrule
\multicolumn{4}{l}{\textit{Reference-set validity}} \\
Rules sufficient             & 96.2 & 89.9 & 91.2 \\
Missing rule ($\geq 2$ raters)    & 0.0 & 0.0 & 0.0 \\
Over-inclusion ($\geq 2$ raters)  & 0.0 & 25.5 & 20.0 \\

\bottomrule
\end{tabular}
}
\end{table}

\subsection{Ground-Truth Reference-Set Analysis}
\label{app:gt-refs}

Across the 231 questions, the ground-truth reference sets contain 844 citations with multiplicity (314 unique articles), averaging 3.65 references drawn from 1.39 documents per question. By annotation granularity, 69.2\% of citations are article-level (조), 29.3\% are sub-clause (항/호), and 1.5\% are appendix forms (별표); scoring collapses sub-clause citations to their governing article, the atomic unit of authority. Reference sets are well-grounded in the indexed graph: 96.8\% of citations resolve to an OKG node, with only 3.2\% pointing outside the constructed graph. The reference distribution is sharply concentrated on the operational core of the corpus (Table~\ref{tab:top-anchors}): the single most-cited article is the Cost-Use Standards pre-approval clause (Art.~73, 65 citations), followed by the international-joint-R\&D and fund-management standards, and by the two sanction-side authorities (Enforcement Decree Art.~26 and Cost-Use Standards Art.~83). These dominant anchors correspond directly to the deterministic expert rules R3 (pre-approval) and R4 (sanction exhaustion).

\begin{table}[t]
\centering
\small
\caption{Five most-cited articles in the ground-truth reference sets and their roles under the deterministic reference rules.}
\resizebox{\columnwidth}{!}{
\begin{tabular}{llr}
\toprule
Article (anchor) & Role & Cites \\
\midrule
Cost-Use Standards Art.~73  & Pre-approval (R3)   & 65 \\
Cost-Use Standards Art.~28  & Joint-R\&D funds    & 40 \\
Cost-Use Standards Art.~13  & Fund usage          & 33 \\
Enforcement Decree Art.~26  & Settlement (R4)     & 31 \\
Cost-Use Standards Art.~83  & Disqualification (R4)& 30 \\
\bottomrule
\end{tabular}
}
\label{tab:top-anchors}
\end{table}

\subsection{Validating the Axis-Decoupling Principle}
\label{app:aug_valid}

The augmentation treats substantive question type and structural difficulty as separate construction axes. Depending on the anchor and injected facets, the same question type can span multiple difficulty levels. Table~\ref{tab:diff-x-qtype} reports the empirical type$\times$difficulty cross-tabulation, which confirms the intended spread: single-clause questions concentrate at L1, exception-paired questions at L2, while the multi-facet and general-principle types populate the L3/L4 tiers where deep reference closures are required. Before filtering, 69.4\% of generated candidates realized their targeted difficulty level; candidates with divergent realized difficulty were discarded from the released set. The facet injection that drives this spread draws on six facet dimensions (transaction, actor-role, recipient-role, geography, lifecycle, and domain) and 14 four-institution parallel-group topics, providing the combinatorial breadth needed to decouple intent from structural depth.

\begin{table}[t]
\centering
\small
\caption{Question-type $\times$ difficulty cross-tabulation, illustrating that the same substantive intent can be realized across multiple difficulty tiers under axis-decoupled augmentation.}
\setlength{\tabcolsep}{4pt}
\begin{tabular}{lrrrrr}
\toprule
Question type & L1 & L2 & L3 & L4 & Total \\
\midrule
Single-clause        & 35 & 13 & 2  & 0  & 50 \\
Exception-paired     & 0  & 26 & 0  & 0  & 26 \\
Institution-specific & 3  & 16 & 0  & 3  & 22 \\
General-principle    & 0  & 0  & 11 & 13 & 24 \\
Multi-facet          & 0  & 0  & 23 & 30 & 53 \\
IITP seed (human)    & 10 & 26 & 9  & 11 & 56 \\
\midrule
Total                & 48 & 81 & 45 & 57 & 231 \\
\bottomrule
\end{tabular}
\label{tab:diff-x-qtype}
\end{table}

\begin{table}
\centering
\caption{Synthetic (LLM-generated) vs.\ human (IITP QA board) splits of RegOps-Bench.}
\label{tab:aug_validation}
\resizebox{\columnwidth}{!}{%
\begin{tabular}{l l rrrr r}
\toprule
& \textbf{Source} & \textbf{L1} & \textbf{L2} & \textbf{L3} & \textbf{L4} & \textbf{All} \\
\midrule
\multicolumn{7}{l}{\textit{(a) Query distribution}} \\
\# Queries & LLM-Generated & 38 & 55 & 36 & 46 & 175 \\
           & IITP-Board    & 10 & 26 &  9 & 11 &  56 \\
Share (\%) & LLM-Generated & 21.7 & 31.4 & 20.6 & 26.3 & 100.0 \\
           & IITP-Board    & 17.9 & 46.4 & 16.1 & 19.6 & 100.0 \\
\midrule
\multicolumn{7}{l}{\textit{(b) Corpus characteristics by difficulty}} \\
Mean hop count      & LLM-Generated & 1.00 & 1.62 & 3.6 & 3.3 & 2.29 \\
                    & IITP-Board    & 1.00 & 1.04 & 1.33 & 2.73 & 1.41 \\
Cross-document (\%) & LLM-Generated &  0.0 &  0.0 & 66.7 & 69.6 & 32.0 \\
                    & IITP-Board    &  0.0 &  0.0 & 11.1 & 54.5 & 12.5 \\
Mean \#GT refs      & LLM-Generated & 1.29 & 2.42 & 6.4 & 6.7 & 4.1 \\
                    & IITP-Board    & 1.40 & 1.69 & 2.89 & 3.64 & 2.21 \\ \midrule
\multicolumn{7}{l}{\textit{(c) RefWalk-35B Claim-F1}} \\
Claim-F1    & LLM-Generated & 52.6 & 57.2 & 28.7 & 31.6 & 43.6 \\
            & IITP-Board    & 36.7 & 30.1 & 30.8 & 39.0 & 33.1 \\
\midrule
\multicolumn{7}{l}{\textit{(d) RefWalk-35B Citation-F1}} \\
Citation-F1 & LLM-Generated & 87.3 & 71.6 & 29.0 & 30.3 & 55.4 \\
            & IITP-Board    & 47.3 & 56.3 & 51.9 & 53.2 & 53.4 \\
\bottomrule
\end{tabular}%
}
\end{table}

Synthetic queries exhibit deeper reference structures than the human board, with more hops, larger reference sets, and a higher cross-document rate. RefWalk achieves similar overall Citation F1 across the two sources, whereas its performance is lower on synthetic L3/L4 questions. These results demonstrate that the augmentation is structurally non-trivial, while the human validation in Table~\ref{tab:human-validation} provides evidence of its practical fidelity.

\subsection{Computational Cost}

\begin{table}[t]
\centering
\footnotesize
\caption{Integrated mean query latency (at 35B backbone) and index-build cost on RegOps-Bench. All latency metrics are reported in mean wall-clock seconds per query. Bold text indicates the best performance in each column.}
\label{tab:integrated_latency}
\resizebox{\columnwidth}{!}{
\begin{tabular}{lrrr}
\toprule
& \multicolumn{2}{c}{\textbf{Query 35B Latency (s)}} & \textbf{Index} \\
\cmidrule(lr){2-3}
\textbf{Method} & \textbf{Retrieval} & \textbf{RAG} & \textbf{LLM calls} \\
\midrule
NativeRAG & \textbf{0.02} & \textbf{7.46} & -- \\
PIKE-RAG          & 0.33          & 24.4          & 718 \\
LightRAG          & 1.46          & 8.48          & 2{,}329 \\
HippoRAG-2        & 2.36          & 9.26           & 1{,}436 \\
Query Decomp      & 4.52          & --           & -- \\ \midrule
RefWalk (Ours) & 4.2    & 9.4           & -- \\
\bottomrule
\end{tabular}
}
\end{table}

In Table~\ref{tab:integrated_latency}, metrics were evaluated on the RegOps dataset ($n = 231$) under identical hardware configurations using a 35B backbone. At retrieval time, RefWalk incurs an overhead of +4.18~s compared to the fastest baseline (NativeRAG). This represents the marginal cost of executing the cross-encoder across three distinct views ($q_{\text{narrow}}$, $q_{\text{mid}}$, $q_{\text{wide}}$) alongside a 1-hop OKG expansion. Notably, RefWalk is $\sim$2.6$\times$ faster than PIKE-RAG in RAG latency (9.4~s vs. 24.4~s). On the indexing side, RefWalk avoids generative LLM calls during index construction, whereas other graph-based methods incur substantial pre-computation overhead.

\subsection{Sensitivity Analysis of the RRM Smoothing Constant $k$}
\label{appendix:rrm_robustness}

\begin{table}[htbp]
\centering
\small
\setlength{\tabcolsep}{5pt}
\caption{Sensitivity analysis of RefWalk with respect to the RRM smoothing constant $k$ on the RegOps dataset ($n=231$, retrieval-only).}
\label{tab:rrm-sensitivity}
\begin{tabular}{r cc cccc}
\toprule
& \multicolumn{2}{c}{\textbf{Overall}} & \multicolumn{4}{c}{\textbf{Recall@10 by Difficulty}} \\
\cmidrule(lr){2-3} \cmidrule(lr){4-7}
$k$ & R@10 & FC@10 & L1 & L2 & L3 & L4 \\
\midrule
10  & 63.51 & 45.89 & 90.62 & 72.63 & 44.00 & 43.10 \\
30  & 63.80 & 45.89 & 90.62 & 73.25 & 44.00 & 43.40 \\
\textbf{60} & 63.80 & 45.89 & 90.62 & 73.25 & 44.00 & 43.40 \\
100 & 63.80 & 45.89 & 90.62 & 73.25 & 44.00 & 43.40 \\
200 & 63.80 & 45.89 & 90.62 & 73.25 & 44.00 & 43.40 \\
\bottomrule
\end{tabular}
\end{table}

As detailed in Table~\ref{tab:rrm-sensitivity}, overall Recall@10 varies by only 0.29 percentage points across a multi-order sweep of $k \in [10, 200]$, and is identical for $k\ge 30$.

\subsection{Citation Granularity}
\label{app:cit}
\begin{table}[t]
\centering
\small
\setlength{\tabcolsep}{3.5pt}
\caption{Citation F1 under different citation granularities on RegOps-Bench (Qwen3.6-35B). The paragraph/item subset contains the 154 questions with at least one paragraph/item-level gold reference.}
\label{tab:gran}
\resizebox{\columnwidth}{!}{
\begin{tabular}{lcccc}
\toprule
\textbf{System}
& \multicolumn{1}{c}{\textbf{Article-level}}
& \multicolumn{1}{c}{\textbf{Exact}}
& \multicolumn{2}{c}{\textbf{Paragraph/Item}} \\
& \textbf{roll-up}
& \textbf{granularity}
& \textbf{subset}
& \textbf{only} \\
\midrule
NativeRAG & 46.7 & 29.0 & 26.1 & 25.6 \\
LightRAG & 40.7 & 27.7 & 23.2 & 23.3 \\
HippoRAG-2 & \underline{47.8} & \underline{29.9} & \underline{27.0} & \underline{29.7} \\
PIKE-RAG & 44.3 & 28.2 & 19.7 & 12.2 \\
\midrule
RefWalk (ours) & \textbf{54.9} & \textbf{38.1} & \textbf{36.2} & \textbf{42.7} \\
\bottomrule
\end{tabular}
}
\end{table}

The main evaluation rolls paragraph/item-level references up to their article-level ancestors, measuring whether a system identifies the correct article without requiring the specific paragraph or item. We therefore additionally evaluate citations at their exact granularity, both over the full benchmark and over questions containing paragraph/item-level references, as well as under paragraph/item-only matching (Table~\ref{tab:gran}). RefWalk consistently performs best across all settings, with a larger margin under finer-grained evaluation. In contrast, several baselines that perform competitively at the article level degrade when paragraph/item-level grounding is required, showing that identifying the correct article does not necessarily entail locating the specific provision that supports the answer.

\subsection{Prompt Template for RefWalk}

\paragraph{Cross-Encoder Reranking.} For the reranking stage, each query-document pair $(q, d)$ is evaluated using a cross-encoder model to compute a deterministic relevance score. The model is optimized via a binary classification objective, prompting it to generate a strict \texttt{yes} or \texttt{no} tokens indicating whether the document satisfies the query. This prompt structure is uniformly applied across all three retrieval views within the RRM-fusion pipeline and is consistently exercised throughout all experiments in this study. The unified ChatML prompt template, which embeds the structural instruction, query placeholder, and target document context, is illustrated in Figure~\ref{fig:reranker_prompt}.

\begin{figure}[htbp]
\centering
\begin{tcolorbox}[
  colback=gray!2!white,
  colframe=black,
  title={Prompt for Cross-Encoder Reranker},
  fontupper=\small\ttfamily,
  arc=1mm,
  boxrule=0.8pt,
  left=3mm, right=3mm, top=2.5mm, bottom=2.5mm
]
\textless\textbar{}im\_start\textbar\textgreater{}system \\
Judge whether the Document meets the requirements based on the Query and the Instruct provided. Note that the answer can only be ``yes'' or ``no''.\textless\textbar{}im\_end\textbar\textgreater{} \\
\textless\textbar{}im\_start\textbar\textgreater{}user \\
\textless{}Instruct\textgreater{}: Given a user question, retrieve legal clauses or regulations that answer the question. \\
\textless{}Query\textgreater{}: \{query\} \\
\textless{}Document\textgreater{}: \{doc\}\textless\textbar{}im\_end\textbar\textgreater{} \\
\textless\textbar{}im\_start\textbar\textgreater{}assistant
\end{tcolorbox}
\caption{The unified ChatML prompt structure used for binary relevance scoring in the cross-encoder reranking stage.}
\label{fig:reranker_prompt}
\end{figure}

\begin{figure*}[t]
\begin{tcolorbox}[
  colback=lightBackground,
  colframe=black,
  fontupper=\small\ttfamily,
  title={Prompt for Topic Anchoring},
  left=4mm, right=4mm, top=3mm, bottom=3mm,
  arc=1mm,
  boxrule=0.8pt
]
Extract a procedural topic and structured conditions from a given question for \textcolor{blue}{\{domain\}}.\\
\\
\#\# Topic\\
- Output a phrase (15 under syllables)\\
- Express a generalized procedural category that matches chapter/article granularity in the regulation corpus\\
- Include the cost category, item type, or document type implicit in the question\\
- Exclude specific numbers, amounts, or actor identities (these belong in conditions)\\
\\
\#\# Conditions\\
For each of the four dimensions below, extract the value mentioned in the question. If a dimension is not mentioned, use "unspecified".\\
\\
- Actor: the top entities whose institutional or organizational type determine which rules apply (e.g., institution type, role, position)\\
- Magnitude: a quantitative threshold or scale that triggers different rules (e.g., monetary amount, count, percentage; keep specific numbers)\\
- Temporal: a time point, period, or sequence relative to the action (e.g., before/after action, deadline, period, frequency, duration)\\
- Situational: any contextual condition not captured by the other dimensions that affects rule applicability (e.g., funding source, collaboration arrangement, employment status, planning status, organizational relation, document type)\\
\\
\#\# Output format\\
Use JSON format. Topic and condition values in \textcolor{blue}{\{language\}}. The parsed output should allow reconstruction of the original question's meaning.\\
\\
\#\# Examples\\
\\
\#\#\# Example 1\\
Question: Can a covered entity share a patient's medical records with an external researcher who is not affiliated with the entity?\\
\{\\
\text{ } "topic": "Disclosure of medical records to external researcher",\\
\text{ } "actor": "Covered entity",\\
\text{ } "magnitude": "unspecified",\\
\text{ } "temporal": "unspecified",\\
\text{ } "situational": "External researcher (not affiliated)"\\
\}\\
\\
\#\#\# Example 2\\
Question: Can I pay the labor cost in cash to a new recruiter at a for-profit organization?\\
\{\\
\text{ } "topic": "Payment of cash labor costs",\\
\text{ } "actor": "For-profit organization",\\
\text{ } "magnitude": "unspecified",\\
\text{ } "temporal": "unspecified",\\
\text{ } "situational": "new recruit"\\
\}\\
\\
\#\#\# Example 3\\
Question: 3천만원 이상 GPU를 구매할 때 사전승인이 필요한가요?\\
\{\\
\text{ } "topic": "장비 구매 사전승인",\\
\text{ } "actor": "unspecified",\\
\text{ } "magnitude": "3천만원 이상",\\
\text{ } "temporal": "구매 전",\\
\text{ } "situational": "unspecified"\\
\}
\end{tcolorbox}
\caption{Generalized prompt structure for procedural topic extraction and anchoring.}
\label{fig:topic}
\end{figure*}

\paragraph{Topic Anchoring $\tau(q)$.} The detailed prompt template for this task is illustrated in Figure~\ref{fig:topic}. The structured topic + facet extraction prompt used by the frozen LLM call $\tau(q)$ produces the Topic Anchor \texttt{topic}, \texttt{actor}, \texttt{temporal}, \texttt{magnitude}, \texttt{situational} for every RegOps and HIPAA query. The \texttt{\{domain\}} placeholder is ``Korean R\&D funding regulations'' for RegOps and ``U.S. healthcare privacy regulations (HIPAA)'' for HIPAA; \texttt{\{language\}} is ``Korean'' or ``English'' respectively. The output is constrained to the \texttt{QueryAnalysis} JSON schema.

\begin{table}[h]
\small
\centering
\caption{Multi-View Query Format structure.}
\label{tab:query_format}
\begin{tabular}{@{}l p{5.5cm}@{}}
\toprule
\textbf{View} & \textbf{Format} \\
\midrule
$q_{\text{narrow}}$ & the original question text. \\
\addlinespace
$q_{\text{mid}}$ & \texttt{[TOPIC]} $\langle$topic$\rangle$ \texttt{[Q]} $\langle$question$\rangle$ \newline 
\texttt{Conditions:} [ACTOR] $\langle$a$\rangle$, [TEMPORAL] $\langle$t$\rangle$, [MAGNITUDE] $\langle$m$\rangle$, [SITUATIONAL] $\langle$s$\rangle$ \\
\addlinespace
$q_{\text{wide}}$ & \texttt{[TOPIC]} $\langle$topic$\rangle$ \newline 
\texttt{Conditions:} [ACTOR] $\langle$a$\rangle$, [TEMPORAL] $\langle$t$\rangle$, [MAGNITUDE] $\langle$m$\rangle$, [SITUATIONAL] $\langle$s$\rangle$ \\
\bottomrule
\end{tabular}
\end{table}

\paragraph{Multi-View Tagged-Query Rendering.} Given the topic anchor produced by $\tau(q)$, the three retrieval views $q_{\text{narrow}}$, $q_{\text{mid}}$, and $q_{\text{wide}}$ are deterministically generated as shown in Table~\ref{tab:query_format}. For instance, the resulting rendered $q_{\text{mid}}$ view is structured as follows:

\begin{center}
\begin{minipage}{1.0\linewidth}
\begin{tcolorbox}[
  colback=gray!5!white, 
  colframe=gray!50!black, 
  arc=1mm, 
  boxsep=3pt, left=4pt, right=4pt, top=4pt, bottom=4pt,
  fontupper=\small\ttfamily
]
[TOPIC] 장비 구매 사전승인 \\
{[Q]} 3천만원 이상 GPU를 구매할 때 사전승인이 \\
\text{ } 필요한가요? \\
Conditions: [ACTOR] all, [TEMPORAL] 구매 전, \\
\text{ } [MAGNITUDE] 3천만원 이상, \\
\text{ } [SITUATIONAL] unspecified
\end{tcolorbox}
\end{minipage}
\end{center}

\paragraph{Schema-Constrained Generation.} This section details the system and user prompt templates employed by the RefWalk framework to realize the schema-constrained attribution mechanism described in \S4.2. Rather than generating free-form prose followed by loose citation footers, the generative model is strictly constrained via a strict JSON schema. The keys of the emitted JSON object must correspond exactly to the \texttt{node\_id} elements present within the retrieved context, and the values are restricted to arrays of granular claims bounded by paragraph-level annotations (e.g., \texttt{[제O항]}). 

This architectural constraint structurally prevents post-hoc rationalization (i.e., ``writing first, citing later''). This setup corresponds directly to the \textit{w/ schema} RefWalk variant evaluated in Tables~\ref{tab:rag} and~\ref{tab:HIPAA} of the main paper. The formalized Korean variant of the system prompt and user template for the RegOps domain are presented in Figure~\ref{fig:refwalk_prompt}.

\begin{figure*}[htbp]
\centering
\begin{tcolorbox}[
  colback=gray!2!white,
  colframe=black,
  title={RefWalk System Prompt Template (RegOps-Bench, Korean)},
  fontupper=\small\ttfamily,
  arc=1mm, boxrule=0.8pt, left=3mm, right=3mm, top=2.5mm, bottom=2.5mm
]
당신은 국가연구개발사업 및 관련 법령/규정 해석을 지원하는 전문가입니다. 사용자가 제공하는 문맥(Context), 질문(Question), 그리고 질문의 주요 조건(주체, 시점, 규모, 상황)을 철저히 분석하여 정확한 답변을 제공해야 합니다. \\
\\
\textbf{\# Instruction} \\
1. 문맥 의존성: 모든 답변 및 근거는 반드시 제공된 'Context' 내에서만 추출해야 합니다. 제공된 문서에 없는 내용을 임의로 지어내거나 외부 지식을 개입하지 마세요. \\
2. 참조 단위 통일(조 단위): JSON의 Key 값은 Context에 명시된 문서의 \texttt{node\_id} (예: ``국가연구개발사업\_연구개발비\_사용기준\_제22조'') 형식과 동일하게 작성해야 합니다. 절대 임의로 Key의 이름을 지어내거나 변형하지 마세요. \\
3. 근거(Claim) 추출 및 세부 단위 표기: 각 참조 조항에서 질문에 답변하기 위해 필요한 구체적인 사실이나 규정 내용을 추출하여 배열(Array) 형태로 나열하세요. 이때 실제 근거가 되는 세부 단위(항, 호)가 존재한다면 내용 앞에 대괄호(예: [제O항])로 표기하세요. \\
4. 예외 및 단서 조항 필수 반영: 제공된 조건(주체, 시점, 규모, 상황)의 값이 'all'(특정되지 않음)이거나 포괄적인 경우, 존재하는 예외 조건이나 단서 조항(예: ``다만, $\sim$'', ``$\sim$의 경우 예외로 한다'', ``중앙행정기관의 장이 인정하는 경우'' 등)을 반드시 탐색하여 근거(Claim)에 포함시켜야 합니다. \\
5. 최종 답변(Answer) 작성: \\
\text{ } - 추출한 근거(Claim)와 예외 조항을 종합하여 질문에 부합하는 명확한 답변을 작성하세요. \\
\text{ } - 예외 조건이 있는 경우, 답변 내에 ``다만, [예외 조건]의 경우 [예외 결과]가 가능합니다/제한됩니다.''의 형태로 명시적으로 고지하세요. \\
\text{ } - 답변 내에 어떤 법령의 몇 조 몇 항에 따른 것인지 명확히 언급하여 신뢰성과 전문성을 높이세요. \\
6. 출력 형식: 반드시 아래의 JSON format으로만 응답해야 하며, JSON 외의 다른 인사말이나 부연 설명은 절대 포함하지 마세요. \\
\\
\textbf{\# Output format (JSON)} \\
\{ \\
\text{ } ``OOO\_제O조'': [ \\
\text{ } \text{ } ``[제1항] 해당 조항에서 도출한 원칙적인 근거 원문 및 내용'', \\
\text{ } \text{ } ``[제2항] 대상별로 다른 규정이 적용되는 경우의 또 다른 원문 내용'', \\
\text{ } \text{ } ``[예외] 해당 조항에 존재하는 예외 및 단서 조항 내용'' \\
\text{ } ], \\
\text{ } ``answer'': ``원칙적인 규정에 대한 설명. 다만, [예외 조건]에 해당하는 경우 [예외 사항]이 적용됩니다.'' \\
\}
\end{tcolorbox}

\vspace{0.5mm} 

\begin{tcolorbox}[
  colback=gray!2!white,
  colframe=black,
  title={RefWalk User Template (RegOps-Bench, Korean)},
  fontupper=\small\ttfamily,
  arc=1mm, boxrule=0.8pt, left=3mm, right=3mm, top=2.5mm, bottom=2.5mm
]
\#\# Context: \\
\{ctx\} \\
\\
The below question is about \{topic\}. The conditions for the topic are as follows: \\
- 주체: \{actor\} \\
- 시점: \{temporal\} \\
- 규모: \{magnitude\} \\
- 상황: \{situational\} \\
\\
\#\# Question: \{q\}
\end{tcolorbox}
\caption{The separated System Prompt and User Template structures used by RefWalk (RegOps-Bench).}
\label{fig:refwalk_prompt}
\end{figure*}

\begin{figure*}[htbp]
\centering
\begin{tcolorbox}[
  colback=gray!2!white,
  colframe=black,
  title={RefWalk System Prompt Template (RegOps, English)},
  fontupper=\small\ttfamily,
  arc=1mm, boxrule=0.8pt, left=3mm, right=3mm, top=2.5mm, bottom=2.5mm
]
You are an expert who supports the interpretation of national research and development (R\&D) projects and related statutes/regulations. You must thoroughly analyze the context (Context), question (Question), and the key conditions of the question (subject, timing, scale, situation) provided by the user to deliver an accurate answer. \\
\\
\textbf{\# Instruction} \\

1. Context Dependency: All answers and grounds must be extracted strictly from within the provided 'Context'. Do not fabricate any content not present in the provided document or introduce external knowledge. \\
2. Uniformity of Reference Unit (Article level): The key values of the JSON must be written in the exact same format as the \texttt{node\_id} specified in the Context (e.g., ``국가연구개발사업\_연구개발비\_사용기준\_제22조''). Never arbitrarily invent or modify the names of the keys. \\
3. Ground (Claim) Extraction and Detailed Unit Notation: Extract the specific facts or regulatory content required to answer the question from each referenced clause, and list them in an array format. If a detailed unit (Paragraph, Subparagraph) that serves as the actual ground exists, indicate it with square brackets (e.g., [제O항]) before the content. \\
4. Mandatory Reflection of Exceptions and Provisos: If the value of the provided conditions (subject, timing, scale, situation) is 'all' (unspecified) or comprehensive, you must search for existing exceptional conditions or provisos (e.g., ``다만, $\sim$'' [provided that, $\sim$], ``$\sim$의 경우 예외로 한다'' [except in the case of $\sim$], ``중앙행정기관의 장이 인정하는 경우'' [cases recognized by the head of the central administrative agency], etc.) and include them in the grounds (Claim). \\
5. Drafting the Final Answer (Answer): \\
\text{ } - Synthesize the extracted grounds (Claim) and exceptions to write a clear answer that aligns with the question. \\
\text{ } - If there is an exceptional condition, explicitly notify it in the answer in the form of ``다만, [예외 조건]의 경우 [예외 결과]가 가능합니다/제한됩니다.'' [Provided that, in the case of [Exceptional Condition], [Exceptional Outcome] is possible/restricted.] \\
\text{ } - Clearly mention which Article and Paragraph of which statute the answer is based on to enhance credibility and expertise. \\
6. Output Format: You must respond strictly in the JSON format below, and never include any other greetings or additional explanations outside of the JSON. \\
\\
\textbf{\# Output format (JSON)} \\
\{ \\
\text{ } ``OOO\_Article O'': [ \\
\text{ } \text{ } ``[Paragraph 1] Original text and details of the principle-based ground derived from the relevant clause'', \\
\text{ } \text{ } ``[Paragraph 2] Another original text content in cases where different regulations apply to each subject'', \\
\text{ } \text{ } ``[Exception] Content of exceptions and provisos existing within the relevant clause'' \\
\text{ } ], \\
\text{ } ``answer'': ``Explanation of the principle-based regulation. Provided that, in the case of [Exceptional Condition], [Exceptional Matter] applies.'' \\
\}
\end{tcolorbox}
\caption{The System Prompt Template structures used by RefWalk (RegOps-Bench).}
\label{fig:refwalk_prompt_eng_1}
\end{figure*}


\begin{figure*}[htbp]
\centering
\begin{tcolorbox}[
  colback=gray!2!white,
  colframe=black,
  title={RefWalk System Prompt Template (HIPAA)},
  fontupper=\small\ttfamily,
  arc=1mm, boxrule=0.8pt, left=3mm, right=3mm, top=2.5mm, bottom=2.5mm
]
You are an AI compliance expert in the U.S. healthcare privacy regulations (HIPAA, 45 CFR Parts 160 and 164). \\
\\
\textbf{\# Instruction} \\
1. Context dependency: Every claim and the final answer MUST be drawn from the provided 'Context' only. Do not invoke outside knowledge or fabricate content. \\
2. Reference-key format: Each JSON key MUST be the `node\_id` exactly as it appears between the [ and ] of a Context passage header — without the surrounding brackets. Do not invent, modify, or concatenate identifiers. (Example: header [spans-passthrough-candidates/sent\_0012-0ec05553] → key ``spans-passthrough-candidates/sent\_0012-0ec05553''.) \\
3. Cite minimally: include only the rule(s) that materially support the answer. The default is one rule per question; cite a second only when the answer genuinely depends on a distinct second rule. Do not echo every passage in the Context. \\
4. Claim extraction: Under each cited node\_id, list the specific facts the rule contributes toward the answer (JSON array of one-sentence English strings). Surface an exception or proviso (``except'', ``provided that'', ``unless'') only when the cited rule itself explicitly contains one — do NOT search for exceptions when none are present in the cited text. \\
5. Final 'answer' field: \\
\text{ } - Concise English synthesis grounded in the cited claims. \\
\text{ } - Reference the regulation in-line by its CFR section (e.g., "§164.502(a)(1)") so the reasoning is traceable. \\
\text{ } - State any explicit exception when it applies. \\
6. Output format: Respond with ONLY the JSON object below. No greetings, no preface, no trailing prose. \\
\\
\textbf{\# Output format (JSON)} \\
\{ \\
\text{ } ``<node\_id>'': [ \\
\text{ } \text{ } ``Primary normative content derived from this rule'', \\
\text{ } \text{ } ``Exception or proviso, only when the cited rule explicitly contains one'', \\
\text{ } ], \\
\text{ } ``answer'': ``Final answer text. Reference the regulation in-line (e.g., §164.502(a)) and state any explicit exception when it applies.'' \\
\}
\end{tcolorbox}
\vspace{0.5mm}
\begin{tcolorbox}[
  colback=gray!2!white,
  colframe=black,
  title={RefWalk User Template (HIPAA)},
  fontupper=\small\ttfamily,
  arc=1mm, boxrule=0.8pt, left=3mm, right=3mm, top=2.5mm, bottom=2.5mm
]
\#\# Context: \\
\{ctx\} \\
\\
The question below is about \{topic\}. The conditions for the topic are as follows: \\
- actor: \{actor\} \\
- temporal: \{temporal\} \\
- magnitude: \{magnitude\} \\
- situational: \{situational\} \\
\\
\#\# Question: \{q\}
\end{tcolorbox}
\caption{The separated System Prompt and User Template structures used by RefWalk (HIPAA).}
\label{fig:refwalk_prompt_eng}
\end{figure*}

\paragraph{NativeRAG Free-Form Generation Baseline.} To isolate and evaluate the baseline generation quality under unconstrained conditions, we implement a conventional free-form generation model paired with an appended citation-footer instruction. This setup serves as the generation standard for multiple baselines, including NativeRAG, LightRAG, HippoRAG-2, and PIKE-RAG, as well as the \textit{w/o schema} ablation configuration of RefWalk. The prompt templates for this baseline are detailed in Figure~\ref{fig:nativerag_prompt}.

\begin{figure*}[htbp]
\centering
\begin{tcolorbox}[
  colback=gray!2!white,
  colframe=black,
  title={NativeRAG System Prompt \& Citation Instruction (RegOps-Bench, Korean)},
  fontupper=\small\ttfamily,
  arc=1mm, boxrule=0.8pt, left=3mm, right=3mm, top=2.5mm, bottom=2.5mm
]
당신은 국가연구개발사업 및 관련 법령/규정 해석을 지원하는 전문가입니다. 주어진 조항(법령·시행령·고시·기준)에 근거하여 사용자의 질문에 한국어로 답변하세요. 조항에 명시되지 않은 사항은 추측하지 말고, 부족한 경우 그 사실을 명시하세요. \\
\\
\textbf{\# Citation-Footer Instruction} \\
답변의 마지막 줄에 다음 형식으로 인용한 조항의 \texttt{node\_id} 목록을 출력하라. \\
예: [참조] 국가연구개발사업\_연구개발비\_사용기준\_제48조, 국가연구개발사업\_연구개발비\_사용기준\_제48조\_제8항 \\
\texttt{node\_id}는 위 'Reference Document List'에 표기된 항목을 그대로 사용하라.
\end{tcolorbox}

\vspace{0.2mm}

\begin{tcolorbox}[
  colback=gray!2!white,
  colframe=black,
  title={NativeRAG System Prompt \& Citation Instruction (RegOps-Bench, English)},
  fontupper=\small\ttfamily,
  arc=1mm, boxrule=0.8pt, left=3mm, right=3mm, top=2.5mm, bottom=2.5mm
]
You are an expert who supports the interpretation of national research and development (R\&D) projects and related statutes/regulations. Based on the provided clauses (statutes, enforcement decrees, notifications, standards), answer the user's questions in Korean. Do not speculate on matters not explicitly stated in the clauses; if information is insufficient, clearly state that fact. \\
\\
\textbf{\# Citation-Footer Instruction} \\
On the very last line of your response, output a list of the \texttt{node\_id}s of the cited clauses in the following format. \\
Example: [참조] 국가연구개발사업\_연구개발비\_사용기준\_제48조, 국가연구개발사업\_연구개발비\_사용기준\_제48조\_제8항 \\
For the \texttt{node\_id}, use the exact items as listed in the `Reference Document List' above.
\end{tcolorbox}

\vspace{0.2mm}

\begin{tcolorbox}[
  colback=gray!2!white,
  colframe=black,
  title={NativeRAG System Prompt \& Citation Instruction (HIPAA)},
  fontupper=\small\ttfamily,
  arc=1mm, boxrule=0.8pt, left=3mm, right=3mm, top=2.5mm, bottom=2.5mm
]
You are a legal/compliance assistant well-versed in U.S. healthcare privacy regulations (HIPAA). Answer the user's question in English, grounded strictly in the rules provided below. Do not speculate beyond what the rules state; if the rules are insufficient to answer, say so explicitly. \\
\\
\textbf{\# Citation-Footer Instruction} \\
On the final line of your answer, list the node\_ids of the rules you cited in the format below. \\
Example: [Citations] spans-passthrough-candidates/sent\_0012-0ec05553, \\ spans-passthrough-candidates/sent\_0010-99ea86fe \\
Use node\_ids exactly as they appear in the `Reference Document List' above; do not invent identifiers.
\end{tcolorbox}

\vspace{0.2mm}

\begin{tcolorbox}[
  colback=gray!2!white,
  colframe=black,
  title={NativeRAG User Template (English)},
  fontupper=\small\ttfamily,
  arc=1mm, boxrule=0.8pt, left=3mm, right=3mm, top=2.5mm, bottom=2.5mm
]
Question: \{question\} \\
\\
Reference Document List: \\
{[\{i\}]} [\{role\}] \{node\_id\} \\
\{text\} \\
\\
\{citation\_instruction\}
\end{tcolorbox}
\caption{The free-form generation prompt topology consisting of separate system and user instruction blocks for the baseline frameworks.}
\label{fig:nativerag_prompt}
\end{figure*}

\subsection{Qualitative Examples}
\label{app:qualitative}

To illustrate the operational behavior of RefWalk (\texttt{Qwen3.6-35B}) across different reasoning depths, we present three qualitative case studies representing schema-bound exception surfacing (L2), exhaustive parallel-institution closure (L3), and cross-document delegation-chain traversal (L4). 

In each case, the model is required to emit a structured per-rule attribution object containing governing article identifiers paired with their specific claims, followed by a synthesized final answer. Verbatim Korean originals are provided alongside polished English translations marked as \texttt{[EN]}.

\begin{figure*}[t]
\centering
\begin{minipage}{1.0\linewidth}
\begin{tcolorbox}[
  colback=gray!3!white,
  colframe=gray!40!black,
  arc=1mm,
  boxsep=3pt, left=5pt, right=5pt, top=5pt, bottom=5pt,
  fontupper=\small\ttfamily
]
\textbf{[Question]}\\
{}[KO] 해외 파트너와 공동 수행 중인 과제의 연구개발비 집행을 계획보다 변경해야 할 경우, 별도 인가나 절차가 필요한가요? 또한 원화 대비 외화 가치가 하락해 예산 잔액이 부족해지는 경우에도 동일한 절차를 밟아야 하나요?\\
{}[EN] If we need to modify the R\&D budget execution plan for a project conducted jointly with an overseas partner, is a separate authorization required? Furthermore, must we follow the same procedure if the budget becomes insufficient due to a decrease in foreign currency value against the Korean Won?

\tcbline

\textbf{[Per-rule attribution]}\\
Cost-Use Standards, Art.~28~(\,$\cdots$\_사용기준\_제28조)\\
~~\textbullet~[제5항] 연구개발기관의 장은 국제공동연구개발비를 원래계획과 다르게 변경하여 사용하려는 때에는 제73조제1항제8호에 따라 중앙행정기관의 장의 사전 승인을 거쳐 협약을 변경한 후 해당 국제공동연구개발비를 계상하여야 한다.\\
~~~~[EN] (5) When the head of a research and development institution intends to use international joint R\&D funds differently from the original plan, they shall appropriate the funds only after obtaining prior approval from the head of the central administrative agency and amending the agreement pursuant to Article 73(1)8.\\
~~\textbullet~[예외] 단, 환율의 변동만으로 금액이 달라지는 경우에는 제외한다.\\
~~~~[EN] [Exception] Provided, That this shall not apply where the amount changes solely due to exchange-rate fluctuations.

Cost-Use Standards, Art.~73~(\,$\cdots$\_사용기준\_제73조)\\
~~\textbullet~[제1항제8호] 국제공동연구개발비를 원래계획과 다르게 변경하려는 경우(단, 환율의 변동만으로 금액이 달라지는 경우에는 제외한다)는 사전 승인 대상이다.\\
~~~~[EN] (1)8. Any modification to international joint R\&D funds deviating from the original plan is subject to prior approval (excluding cases where the amount changes solely due to exchange-rate fluctuations).

\tcbline

\textbf{[Answer]}\\
{}[KO] 해외 파트너와 공동 수행 중인 과제(국제공동연구개발사업)의 연구개발비 집행을 계획보다 변경해야 할 경우, 원칙적으로 중앙행정기관의 장의 사전 승인을 받아 협약을 변경한 후 계상해야 합니다(사용기준 제28조제5항 및 제73조제1항제8호). 다만, 원화 대비 외화 가치 하락으로 예산 잔액이 부족해지는 등 `환율의 변동만으로 금액이 달라지는 경우'에는 별도의 사전 승인 절차가 면제됩니다.\\
{}[EN] In principle, changing the budget execution for a joint project with an overseas partner (international joint R\&D) requires prior approval from the head of the central administrative agency and a subsequent agreement amendment before appropriation (Art. 28(5) and Art. 73(1)8 of the Cost-Use Standards). However, if the budget shortfall stems solely from exchange-rate fluctuations, such as a depreciation of the foreign currency against the Won, the requirement for a separate prior approval is waived.
\end{tcolorbox}
\end{minipage}
\caption{Qualitative example of schema-bound exception surfacing (L2). The model correctly isolates the tail exception clause and binds it to its parent statutory anchor instead of emitting a free-form textual caveat.}
\label{fig:ex1_l2}
\end{figure*}

\noindent\textbf{Analysis of Example 1.} As illustrated in Figure~\ref{fig:ex1_l2}, the ground-truth rule closure requires targeting exactly $\{\text{Art.~28, Art.~73}\}$. RefWalk achieves perfect citation precision and high claim alignment (citation F1$ = 1.00$, claim F1 $= 0.89$). Crucially, rather than treating the exchange-rate clause as an unstructured text block, the attribution schema structurally binds the exception directly to its governing statutory articles, ensuring that the final counterfactual synthesis is grounded in traceable textual support.

\begin{figure*}[t]
\centering
\begin{minipage}{1.0\linewidth}
\begin{tcolorbox}[
  colback=gray!3!white,
  colframe=gray!40!black,
  arc=1mm,
  boxsep=3pt, left=5pt, right=5pt, top=5pt, bottom=5pt,
  fontupper=\small\ttfamily
]
\textbf{[Question]}\\
{}[KO] 저희 기관에서 올해 사업비 집행 중 직접비가 부족하여 간접비 조정을 받았고, 이로 인해 간접비 잔액이 늘었습니다. 남은 잔액으로 연구개발능률성과급을 지급하려 하는데, 성과급으로 계상할 수 있는 상한 비율이 어떻게 되며 주의할 점이 있을까요?\\
{}[EN] Our institution received an indirect-cost adjustment this year due to a shortfall in direct costs, which consequently increased our remaining indirect-cost balance. We plan to pay R\&D efficiency performance bonuses from this surplus. What is the maximum allowable ratio for this bonus, and are there any critical compliance issues we should note?

\tcbline

\textbf{[Per-rule attribution]}\\
Art.~43~(Government-Funded Institute)\\
~~\textbullet~[제6항] 정부출연기관의 장은 연구개발능률성과급을 해당 기관의 회계연도 간접비 총액의 10퍼센트 이하로 계상하여야 한다.\\
~~~~[EN] (6) The head of a government-funded research institute shall budget R\&D efficiency performance bonuses at no more than 10\% of the institution's total indirect costs for the relevant fiscal year.\\
~~\textbullet~[준용] 제43조는 정부출연기관에 대한 규정이나, 대학 및 기타비영리기관에 대해 준용됨(제51조제4항, 제59조).\\
~~~~[EN] [Mutatis Mutandis] Although Article 43 explicitly addresses government-funded institutes, it applies mutatis mutandis to universities and other non-profit organizations via Articles 51(4) and 59.

Art.~51~(University)\\
~~\textbullet~[제4항] 대학이 사용하는 직접비의 부족에 따른 간접비 조정 $\cdots$ 및 연구개발능률성과급 계상에 관하여는 제43조제3항, 제4항, 제6항을 준용한다. 이 경우 ``정부출연기관''은 ``대학''으로 본다.\\
~~~~[EN] (4) Regarding indirect-cost adjustments due to direct-cost shortfalls $\cdots$ and the budgeting of R\&D efficiency performance bonuses for universities, Article 43(3), (4), and (6) shall apply mutatis mutandis. In such cases, ``government-funded research institute'' shall be read as ``university.''

Art.~59~(Other Non-Profit)\\
~~\textbullet~[본문] 기타비영리기관이 사용하는 간접비의 사용기준에 관하여는 제43조를 준용한다. 이 경우 ``정부출연기관''은 ``기타비영리기관''으로 본다.\\
~~~~[EN] [Main Text] The standards for using indirect costs by other non-profit organizations shall be governed mutatis mutandis by Article 43. In such cases, ``government-funded research institute'' shall be read as ``other non-profit organization.''

Art.~69~(For-Profit)\\
~~\textbullet~[제3항] 영리기관의 장은 연구개발능률성과급을 해당 연구개발과제 간접비의 10퍼센트 이하로 계상하여야 한다.\\
~~~~[EN] (3) The head of a for-profit institution shall budget R\&D efficiency performance bonuses at no more than 10\% of the indirect costs allocated to the relevant R\&D project.

\tcbline

\textbf{[Answer]}\\
{}[KO] 연구개발능률성과급의 상한은 기관 유형에 따라 적용 기준이 다릅니다. 첫째, 정부출연기관·대학·기타비영리기관은 제43조(대학은 제51조제4항, 기타비영리기관은 제59조 준용)에 따라 `해당 기관의 회계연도 간접비 총액'의 10퍼센트 이하로 계상해야 합니다. 즉, 개별 과제별 잔액이 아닌 기관 전체의 간접비 총액이 기준입니다. 둘째, 영리기관은 제69조제3항에 따라 `해당 연구개발과제 간접비'의 10퍼센트 이하로 계상해야 합니다. 간접비 조정으로 잔액이 늘었더라도 위 10\% 상한을 초과하여 계상·지급할 수 없습니다.\\
{}[EN] The ceiling for R\&D efficiency performance bonuses varies by institution type. First, for government-funded institutes, universities, and other non-profits (governed by Art. 43; applied mutatis mutandis to universities via Art. 51(4) and to non-profits via Art. 59), the bonus is capped at 10\% of the institution's total indirect costs for the fiscal year. This means the baseline is the aggregate indirect costs of the entire institution, not the individual project balance. Second, for-profit institutions must cap the bonus at 10\% of the relevant project's specific indirect costs pursuant to Article 69(3). Even if the indirect-cost balance expands due to adjustments, payments cannot exceed these respective 10\% statutory caps.
\end{tcolorbox}
\end{minipage}
\caption{Qualitative example of exhaustive parallel-institution closure (L3). The model dynamically tracks legal \textit{mutatis mutandis} (\textit{deemed application}) edges across distinct organizational frameworks to construct an aggregate compliance answer.}
\label{fig:ex2_l3}
\end{figure*}

\noindent\textbf{Analysis of Example 2.} As demonstrated in Figure~\ref{fig:ex2_l3}, this instance evaluates a multi-institutional parallel rule structure. To construct a legally sound answer, the model must map the entire parallel web without missing the critical distinction between institution-level and project-level caps. RefWalk successfully establishes full reference closure by identifying all four parallel anchors (citation $F_1 = 1.00$). Rather than compressing distinct corporate entities into a single generic rule, it preserves the formal \textit{mutatis mutandis} relationships as discrete, attributable statutory claims.

\begin{figure*}[t]
\centering
\begin{minipage}{1.0\linewidth}
\begin{tcolorbox}[
  colback=gray!3!white,
  colframe=gray!40!black,
  arc=1mm,
  boxsep=3pt, left=5pt, right=5pt, top=5pt, bottom=5pt,
  fontupper=\small\ttfamily
]
\textbf{[Question]}\\
{}[KO] 저희 기관이 해외의 A대학교와 B연구소 두 곳에 공동으로 개발 업무를 위탁하고 있는데, 현지 지급을 위해 기존 예산을 증액하려고 합니다. 이때 해외 공동 수탁처에 대한 자금 이관 및 증빙 자료 관리는 통합계좌와 통합정보시스템 상에서 어떻게 처리해야 혁신법 절차상 무리가 없을까요?\\
{}[EN] Our institution is jointly outsourcing development tasks to two overseas entities (University A and Research Institute B), and we intend to increase the existing budget for local disbursements. To ensure compliance with the procedural requirements of the Innovation Act, how should we handle the transfer of funds and the management of supporting documentation for these overseas co-trustees within the integrated account and the integrated information system?

\tcbline

\textbf{[Per-rule attribution]}\\
Enforcement Decree, Art.~24~(Tier 2)\\
~~\textbullet~[제1항] 연구개발기관의 장은 $\cdots$ 통합정보시스템에서 연구개발과제마다 별도의 계정을 생성하여 $\cdots$ 연구개발비카드를 발급받아야 한다.\\
~~~~[EN] (1) The head of a research and development institution shall establish a separate account for each R\&D project within the integrated information system and be issued an R\&D funds card $\cdots$\\
~~\textbullet~[제4항] 연구개발기관의 장은 $\cdots$ 통합 계좌를 개설하여 연구개발비를 운영할 수 있다. 다만, 제19조제1항제1호, 제2호 또는 제4호에 해당하는 $\cdots$ 경우는 제외한다.\\
~~~~[EN] (4) The head of a research and development institution may open an integrated account to manage R\&D funds; provided, that institutions falling under Article 19(1)1, 2, or 4 shall be excluded.

Cost-Use Standards, Art.~28 Hallmarks~(Tier 4)\\
~~\textbullet~[제3항] 연구개발과제계정에서 국외에 소재한 기관 또는 외국인의 계좌로 이체 $\cdots$ 된 국제공동연구개발비는 사용된 연구개발비로 본다.\\
~~~~[EN] (3) International joint R\&D funds transferred from an R\&D project account to an account of an overseas institution or a foreign national $\cdots$ shall be deemed as expended R\&D funds.\\
~~\textbullet~[제4항] 국외기관 간 계약서, 계좌이체증명 등 증명자료를 갖추어야 한다.\\
~~~~[EN] (4) Supporting documentation, including contracts with the overseas entity and bank transfer confirmations, must be maintained.\\
~~\textbullet~[제5항] 국제공동연구개발비를 원래계획과 다르게 변경하여 사용하려는 때에는 제73조제1항에 따라 사전 승인을 거쳐 협약을 변경한 후 계상하여야 한다.\\
~~~~[EN] (5) When intending to use international joint R\&D funds differently from the original plan, the funds shall be appropriated only after obtaining prior approval and amending the agreement pursuant to Article 73(1).

\tcbline

\textbf{[Answer]}\\
{}[KO] 첫째(예산 증액), 국제공동연구개발비를 증액하려면 사용기준 제28조제5항에 따라 중앙행정기관의 장의 사전 승인을 받아 협약을 변경한 후 계상해야 합니다(단순 환율 변동은 예외). 둘째(자금 이관·회계), 과제계정에서 해외 기관 계좌로 이체된 금액은 제28조제3항에 따라 사용된 연구개발비로 보며, 시행령 제24조제1항에 따라 통합정보시스템에서 과제별 별도 계정을 생성해 관리하고 효율적 관리를 위해 통합 계좌를 운영할 수 있습니다(특정 유형 기관은 제외). 셋째(증빙), 제28조제4항에 따라 계약서·계좌이체증명 등 증명자료를 갖추어야 합니다.\\
{}[EN] First (Budget Increase): To increase international joint R\&D funds, you must obtain prior approval from the head of the central administrative agency and amend the agreement before appropriation pursuant to Article 28(5) of the Cost-Use Standards (except for pure exchange-rate fluctuations). Second (Fund Transfer \& Accounting): Sums transferred from the project account to an overseas institution's account are deemed spent R\&D funds under Article 28(3). In accordance with Article 24(1) of the Enforcement Decree, a separate account per project must be managed within the integrated information system, and an integrated account may be operated for administrative efficiency (excluding certain types of institutions). Third (Evidentiary Proof): You must maintain formal supporting evidence, such as contracts with the foreign entities and bank transfer records, pursuant to Article 28(4).
\end{tcolorbox}
\end{minipage}
\caption{Qualitative example of cross-document chain traversal (L4). The reasoning path spans from high-level statutory frameworks (Tier 2 Enforcement Decree) down to micro-level procedural constraints (Tier 4 Cost-Use Standards).}
\label{fig:ex3_l4}
\end{figure*}

\noindent\textbf{Analysis of Example 3.} As shown in Figure~\ref{fig:ex3_l4}, the legal reasoning path spans two distinct levels of authority, successfully tracing a hierarchical delegation link (\textsc{Delegates\_To}/\textsc{Specifies}) from Enforcement Decree Art.~24 down to the international joint R\&D provisions in Cost-Use Standards Art.~28. RefWalk successfully navigates this cross-document dependency chain without introducing hallucinated citations. 

However, consistent with the precision-first behavior observed in our empirical evaluation, the model selectively recovers the primary structural anchors of the chain while missing surrounding contextual siblings (e.g., Cost-Use Standards Art.~12, 13, and 27). This behavior accounts for its lower article-level citation recall ($0.40$) on this specific problem instance, directly exposing the unresolved recall headroom that \textsc{RegOps-Bench} is tailored to isolate.

\end{document}